\documentclass[10pt,twocolumn,letterpaper]{article}

\usepackage{cvpr}              
\definecolor{cvprblue}{rgb}{0.21,0.49,0.74}
\usepackage[pagebackref,breaklinks,colorlinks,allcolors=cvprblue]{hyperref}
\usepackage{listings}
\usepackage[dvipsnames, table]{xcolor}
\usepackage{soul}
\usepackage{makecell}    
\usepackage{tabularx}
\usepackage{pifont}
\usepackage{cprotect}
\usepackage[skins,breakable]{tcolorbox}
\usepackage{xcolor} 
\usepackage{lipsum}
\usepackage{float}

\newcommand{\cmark}{\textcolor{green!60!black}{\ding{52}}}
\newcommand{\xmark}{\textcolor{red!70!black}{\ding{55}}}
\newcommand{\simbeige}{\textcolor[RGB]{180,120,50}{$\sim$}}

\def\paperID{09310} 
\def\confName{CVPR}
\def\confYear{2026}

\title{\textit{Simple} Agents Outperform \textit{Experts} in Biomedical Imaging \\ Workflow Optimization}

\author{Xuefei (Julie) Wang$^{1}$
\and
Kai A. Horstmann$^{2}$
\and
Ethan Lin$^{2}$
\and
Jonathan Chen$^{2}$
\and
Alexander R. Farhang$^{1}$
\and
Sophia Stiles$^{1}$
\and
Atharva Sehgal$^{3}$
\and
Jonathan Light$^{4}$
\and
David Van Valen$^{1}$
\and
Yisong Yue$^{1}$
\and
Jennifer J. Sun$^{2}$ 
\and
\vspace{0.3em}
{\normalsize $^{1}$Caltech} \quad {\normalsize $^{2}$Cornell} \quad {\normalsize $^{3}$UT Austin} \quad {\normalsize $^{4}$Rensselaer Polytechnic Institute}
}

\begin{document}
\maketitle
\begin{abstract}
Adapting production-level computer vision tools to bespoke scientific datasets is a critical ``last mile'' bottleneck. Current solutions are impractical: fine-tuning requires large annotated datasets scientists often lack, while manual code adaptation costs scientists weeks to months of effort. We consider using AI agents to automate this manual coding, and focus on the open question of optimal agent design for this targeted task. We introduce a systematic evaluation framework for agentic code optimization and use it to study three production-level biomedical imaging pipelines. We demonstrate that a simple agent framework consistently generates adaptation code that outperforms human-expert solutions. Our analysis reveals that common, complex agent architectures are not universally beneficial, leading to a practical roadmap for agent design. We open source our framework and validate our approach by deploying agent-generated functions into a production pipeline, demonstrating a clear pathway for real-world impact. The code can be found here: \href{https://github.com/xuefei-wang/simple-agent-opt}{https://github.com/xuefei-wang/simple-agent-opt}
\end{abstract}    
\section{Introduction}
\label{sec:introduction}

Automated computer vision (CV) tools are rapidly being adopted as production-level solutions in clinical and laboratory settings, fundamentally reshaping scientific discovery in biomedical imaging~\cite{laubscher2024accurate, Stringer2025, medsamNatCom, wang2023scientific}. Despite this progress, a critical and ubiquitous ``last mile'' bottleneck---\textbf{tool adaptation}---still persists. When a scientist applies these tools to their own bespoke datasets, models frequently underperform or fail~\cite{guan2021domain, alfasly2025validation, de2025current, lin2025impact, ma2024multimodality, yan2025categorization}  due to inevitable variability in acquisition conditions between labs, such as different microscopes, lighting, resolutions, staining protocols, or unique artifacts~\cite{trisovicLargescaleStudyResearch2022, leeCellposeChallenge2023, jahanifar2025domain, zhang2022benchmarking, howard2021impact}.

\begin{figure}[t]
  \centering
  \includegraphics[width=1.0\linewidth]{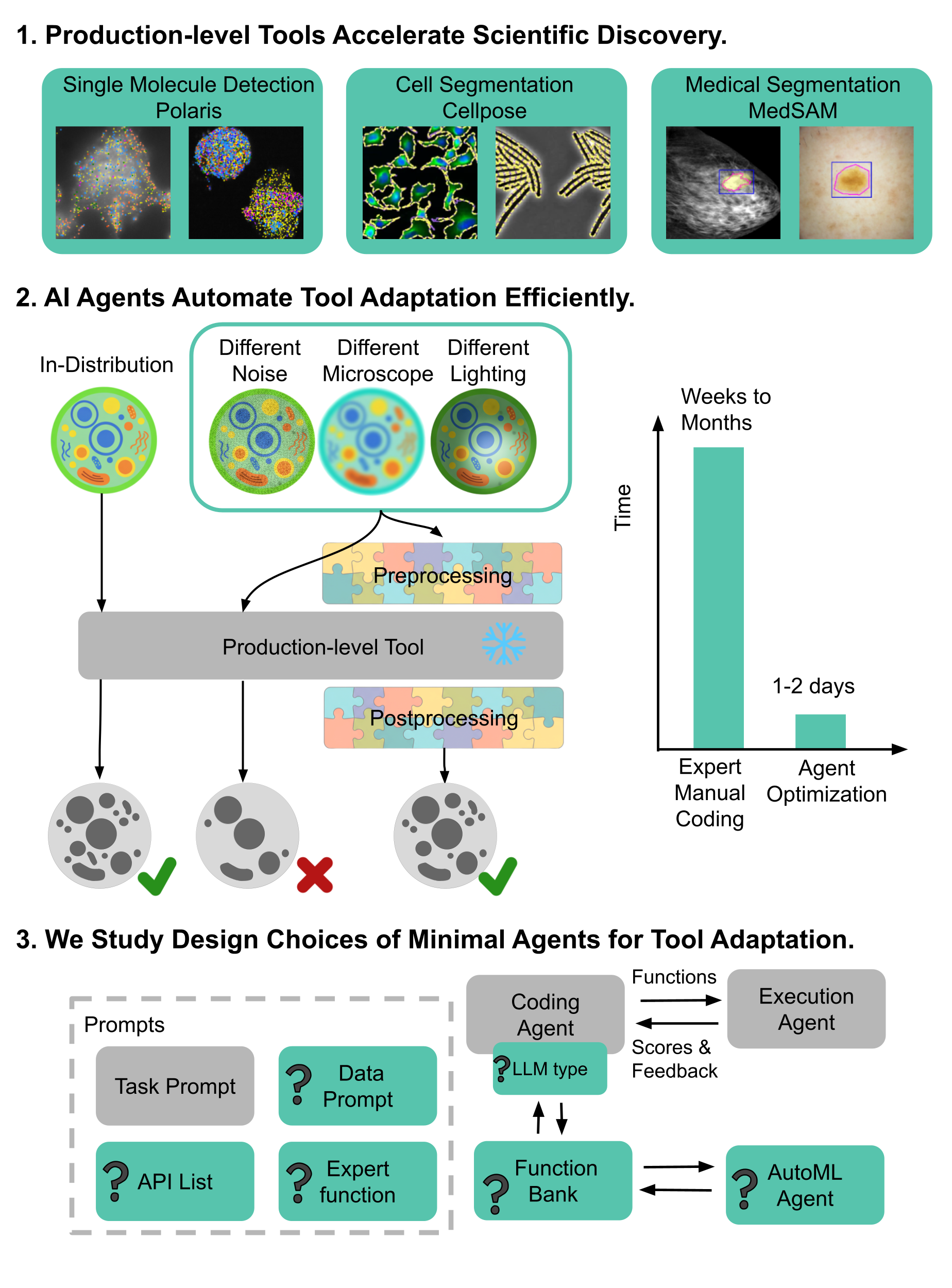}
  \vspace{-2em}
  \caption{Overview. \textit{(Top)} Production-level tools~\cite{laubscher2024accurate, stringer2025cellpose3, medsamNatCom} accelerate scientific discovery but face a ``last mile'' adaptation bottleneck. \textit{(Middle)} Domain experts spend weeks to months manually coding preprocessing and postprocessing steps in order to adapt the tools to their bespoke datasets. AI agents can automate this adaptation, but it remains unclear how to navigate their complex design space to build simple, practical agents. \textit{(Bottom)} Our work systematically studies the design choices of tool adaptation agents.
  }
  \label{fig:overview}
  \vspace{-0.2in}
\end{figure}

Current solutions to this adaptation bottleneck are hindered by the need for either massive amounts of labeled data or prohibitively long manual development cycles. Scientists are typically forced to choose between: (1) Fine-tuning complex models, a process that is data-inefficient and requires a large, annotated training set (e.g., thousands of images) that is often unavailable to individual labs~\cite{litjens2017survey, orouji2024domain, zhang2024low}; or (2) Manually writing custom pre- and post-processing code to bridge the domain gap, which can take a scientist weeks or months, significantly diverting valuable time away from scientific discovery (Figure~\ref{fig:overview}).

Recent work in agentic AI presents a new way to tackle this adaptation bottleneck.
In principle, an AI agent could use the small ``gold-standard'' validation set (typically 10-100 images) that scientists do have as an objective function to automatically generate the necessary adaptation code.

However, most existing AI Agents may not be directly applicable to this specific, highly-demanded application due to their sophisticated architecture or specialized target tasks. ``AI Agents for Science'' are often large, complex systems—featuring hierarchical planning and large tool spaces—designed for high-level, open-ended discovery~\cite{lu2024ai, huang2025biomni, jin2025stella} or specialized scientific tasks~\cite{biodiscoveryagent, wang2025spatialagent, qu2025crispr, li2025co}, rather than the targeted adaptation of existing tools. Concurrently, while MLE agents~\cite{jiang2025aide, liu2025ml, yang2025r} are progressing quickly, they typically focus on building new solutions from scratch, not on integrating with and tuning existing, production-level scientific pipelines. Thus, for the narrower problem of tool adaptation, it remains unclear whether such complex designs are necessary or which specific design components drive performance.

Therefore, we set out to answer: {\textit{\color{blue} What is the most practical and simplest agent framework that can reliably adapt a fixed, pretrained production tool to a new, bespoke dataset?}}

To this end, we introduce a systematic evaluation framework for benchmarking agentic code optimization for tool adaptation. We apply this framework to three production-level biomedical imaging pipelines: Polaris~\cite{laubscher2024accurate}, Cellpose~\cite{stringer2025cellpose3}, and MedSAM~\cite{medsamNatCom}—chosen for their collective coverage of the full spectrum of biological length scales. Our investigation yields a promising and practical path to solving the ``last mile'' tool adaptation problem. To summarize, our key contributions are: 

\begin{itemize}
    \item We demonstrate that a simple agent framework can consistently generate adaptation code that outperforms human-expert-derived solutions---the official production-optimized code from the original tool authors.
    \item Our systematic analysis of the interplay between agent design and the solution space shows that common, complex agent designs are not necessarily beneficial for this focused task, which leads to a practical design roadmap for researchers.
    \item We open-source our framework to facilitate future agent studies and validate the agentic approach by deploying the first agent-generated functions into a production pipeline (anonymous GitHub PR merged into the official codebase, see appendix), demonstrating a pathway for real-world impact.
\end{itemize}


\section{Related Work}
\label{sec:related_work}

\textbf{AI Agents for Science and MLE.}  
In the MLE domain, agentic systems are rapidly increasing in sophistication. Works like AIDE~\cite{jiang2025aide} and others~\cite{liu2025ml, yang2025r} employ complex tree-search strategies for automated model building. Benchmarks~\cite{chan2024mle, huang2023mlagentbench} developed to evaluate these systems often use self-contained, Kaggle-style tasks. This line of work primarily focuses on building complex agents to conduct ML tasks from scratch and thus does not directly address the ``last mile'' adaptation bottleneck for real-life scientific tools.

This trend is mirrored in scientific applications, which fall into two main categories. First, general-purpose agents aim for open-ended discovery~\cite{lu2024ai, huang2025biomni, jin2025stella}, often relying on sophisticated multi-step workflows. Second, specialized agents are custom-built for bespoke tasks like perturbation design~\cite{biodiscoveryagent} or gene-editing automation~\cite{qu2025crispr}. Neither category is suitable for the tool adaptation problem: the former are too general and complex for this targeted task, while the latter are not generalizable solutions for adapting existing pipelines.

Therefore, it remains unclear whether the complex designs from these fields are necessary for the focused and practically important task of tool adaptation. Furthermore, a systematic, component-wise analysis of the agent design space for this task is lacking.

{%
\newcolumntype{C}{>{\centering\arraybackslash}X}
\newcolumntype{M}[1]{>{\centering\arraybackslash}p{#1}}
\newcolumntype{P}[1]{>{\raggedright\arraybackslash}p{#1}}
\renewcommand{\arraystretch}{1.2}
\begin{table*}
  \small
  \centering
  \setlength{\tabcolsep}{1.5pt}
  \caption{Comparison of works most related to our agent study. Symbols: \cmark{} --- criterion met; \xmark{} --- criterion not met; \simbeige{} --- partially met.}
  \label{tab:example}
  \begin{tabularx}{\textwidth}{lP{3.4cm}P{3.15cm}M{2.95cm}CC}
    \toprule
    Work & Primary Goal & Agent frameworks complexity & Exceeds expert baseline (human effort) & Systematic study of agent designs & Production-level deployment \\
    \midrule
    MLAgentBench~\cite{huang2023mlagentbench} & \textbf{Benchmark}: MLE Agents & \textbf{Minimal}: ReAct, code edit + execution & - & \xmark & \xmark \\
    MLE-bench~\cite{chan2024mle} & \textbf{Benchmark}: MLE Agents & \textbf{Heavy}: tree search, terminal access & \makecell{\simbeige\footnotemark[1]\\{\footnotesize(Kagglers weeks-months)}} & \xmark & \xmark \\
    RE-Bench \cite{wijk2024re} & \textbf{Benchmark}: Research Engineering Agents & \textbf{Mixed}: tree search; code edit + execution & \makecell{\xmark\footnotemark[2]\\{\footnotesize(Experts 8-hours)}} & \xmark & \xmark \\
    ScienceAgentBench~\cite{chen2024scienceagentbench} & \textbf{Benchmark}: Scientific Discovery Agents & \textbf{Mixed}: terminal access; code edit + execution & - & \cmark & \xmark \\
    Biomni~\cite{huang2025biomni} & \textbf{Framework}: Scientific Tool Orchestration Agents & \textbf{Heavy}: large toolspace, RAG & - & \xmark & \cmark \\
    \textbf{Ours} & \textbf{Feasibility Study}: Tool Adaptation Agents & \textbf{Minimal}: code edit + execution & \makecell{\cmark\\{\footnotesize(Experts weeks-months)}} & \cmark & \cmark \\
    \bottomrule
  \end{tabularx}
\end{table*}
}%

\noindent\textbf{Low-Data Tool Adaptation.} The scientific adaptation bottleneck is driven by the fact that biological datasets are inherently bespoke and heterogeneous~\cite{trisovicLargescaleStudyResearch2022, leeCellposeChallenge2023, jahanifar2025domain, zhang2022benchmarking, howard2021impact, lin2025impact}. Minor variations in laboratories, equipment, or experimental conditions create unique statistical distributions, leading to domain shift~\cite{orouji2024domain, zhang2024low}. Given that scientists cannot afford to collect and annotate new, large-scale training sets for every experiment, adapting existing tools to new datasets remains an essential, unsolved step in scientific workflows. 

To mitigate this, substantial effort has been dedicated to task-specific solutions, such as customizing additional processing steps~\cite{stringer2025cellpose3}, developing novel networks or training recipes~\cite{schafer2024overcoming, wang2021annotation, zhang2025generative}, or test-time adaptation~\cite{chen2024each, karmanov2024efficient}. However, these approaches are tightly coupled to the task at hand and cannot be easily generalized to different tools or problems. In contrast, our work studies how general, minimal agents can be used for adaptation in this low-data regime.

\noindent\textbf{Classic AutoML.} Automated Machine Learning (AutoML) automates many pipeline stages of method development, including data processing, feature engineering, and neural architecture search \cite{Komer2019, akiba2019Optuna, kanterDeepFeatureSynthesis2015, zimmer-tpami21a, JMLR:v24:20-1355, feurer-arxiv20a, feurer-neurips15a, agtabular}. However, its frameworks are constrained by optimizing a fixed, manually defined search space. This process, as seen in tools like Optuna \cite{akiba2019Optuna}, requires tedious expert effort to curate functions and define search ranges carefully. 
LLM-based agents shift this paradigm. By generating code, they are not limited to a pre-defined set of functions, but can dynamically generate their own parameter and code space. This distinction creates a potential new synergy between agents \& AutoML. An LLM agent can perform the structural search (e.g. identifying which functions to combine), while a classic AutoML component handles the parameter search (e.g., finding the optimal clip limit). Part of our work explores this intersection.

\section{Agent Design Space}
\label{sec:approach}

To systematically evaluate agent designs for tool adaptation, we first decompose the existing agent frameworks into their core modules, establishing a simple ``Base Agent''. We then build upon this foundation by systematically integrating more complex components. This bottom-up framework allows us to test individual design choices in isolation and provide clear insights into their respective impacts.

\footnotetext[1]{Best agent medaled on $\sim$17\% of tasks on the Kaggle leaderboard.}
\footnotetext[2]{Humans consistently lead minimal agent after $\sim$2 hours.}
\addtocounter{footnote}{2}

\subsection{A Base Agent Framework}

The agent system is tasked with iteratively generating code-based solutions (pairs of preprocessing and postprocessing functions), which are then applied to a production-level tool and scored against a small validation set. 

The core structure to achieve this is composed of three essential components (Fig.~\ref{fig:overview}):

\begin{itemize}
    \item \textbf{Task Prompt}: The initial problem specification, which is required for the coding agent to understand the goal and write the desired functions.
    \item \textbf{Coding Agent}: An LLM-based agent whose role is to generate candidate functions.
    \item \textbf{Execution Agent}: A module that receives the generated function, embeds it into the scientific workflow, executes the pipeline, and returns execution feedback and scores back to the coding agent.
\end{itemize}

However, in a specialized scientific domain, this core structure alone may lack the context---such as the nature of the data and available APIs---to generate relevant or functional code. Therefore, we define our practical ``Base Agent'' to include two context components in its prompt, which serves as our baseline for systematic exploration:
\begin{itemize}
    \item \textbf{Data Prompt}: Context on the nature of the data~\cite{chen2024scienceagentbench}.
    \item \textbf{API List}: A list of relevant APIs with docstrings~\cite{wang2024openhands, wang2023voyager}.
\end{itemize}

\subsection{Components of the Agent Design Space}
\label{sec:approach-advanced-components}

Building from the ``Base Agent'', we then explore the agent design space by systematically ablating, adding, and altering more complex components. We focus on components and design choices that are common in the literature or have a significant impact on practicality. First, we conduct an ablation study to confirm the necessity of the ``Data Prompt'' and ``API List''. Then, we identify four key axes of additional augmentation of the agent system (Fig.~\ref{fig:overview}): 

\begin{itemize}
    \item \textbf{LLM Type}: This component defines the model used for the Coding Agent, encompassing a wide range of options, varying in sizes, training focuses, development (open-source vs. closed-source), and providers \cite{chan2024mle, huang2023mlagentbench, chen2024scienceagentbench, toledo2025ai}.
    \item \textbf{Expert Functions}: This component inserts human-expert optimized functions into the prompt to study if they provide guidance and serve as effective in-context examples~\cite{shypula2024learningperformanceimprovingcodeedits}.
    \item \textbf{Function Bank}: This component functions as a persistent memory of its previously generated functions. When enabled, previous functions will be selected and fed back into the agent's prompt to guide further exploration~\cite{wang2023voyager, disciple}.
    \item \textbf{AutoML Agent}: This component adds an explicit hyperparameter search step. When enabled, the agent will be invoked periodically to analyze generated functions from the function bank, identify optimizable parameters, and run a hyperparameter search to fine-tune the parameters~\cite{AutoMLAgentMultiAgentLLMtrirat2025, SequentialLargeLanguagemahammadli2025, laoGptTuner}.
\end{itemize}
\section{Experiment Setup}
\label{sec:experiment_setup}

\subsection{Configurations}

We defined the experimental setup for each design component as follows: 
\begin{itemize}
    \item \textbf{Data Prompt}: We provided biologically relevant information about the datasets (e.g., ``medical,'' ``cell,'' ``fluorescent,'' ``microscopy,'') and image channel interpretations (e.g., ``nucleus'', ``cytoplasm'', ``empty'').
    \item \textbf{API List}: We curated a comprehensive list of 98 relevant functions from the OpenCV, Skimage, and Scipy libraries.
    \item \textbf{LLM Types}: We tested three models of varying capabilities: a large general-purpose model (GPT-4.1), a model noted for its reasoning ability (o3), and a smaller open-source model (Llama 3.3-70B-Instruct-Turbo).
    \item \textbf{Expert Functions}: While the inputs for all tasks are relatively standard images, the output format is highly task-dependent. Therefore, when this component is enabled, we provide the expert-written postprocessing functions in the prompt to serve as in-context examples.
    \item \textbf{Function Bank}: All generated functions and their scores were saved to a function bank. In each subsequent iteration, the top 3 and bottom 3 performing functions were sampled and fed back into the prompt context.
    \item \textbf{AutoML Agent}: The agent was invoked every 5 iterations to select the top-3 functions and optimize each for 24 trials.
\end{itemize}

For each agent configuration, we ran the optimization procedure 20 times with different random seeds. Each run generated 60 trials (20 iterations with 3 function pairs per iteration). To mitigate overfitting, the final performance was not based on the single best-validation function. Instead, we selected the top 15 functions from all 20 runs based on validation scores and reported the maximum test score.

\subsection{Case Studies}

We choose three biomedical imaging applications as case studies to evaluate the design choices. These cases are highly representative as they collectively span the full spectrum of biological length scales, ensuring our findings are broadly applicable:

\begin{center}
\vspace{-0.15in}
\begin{tabular}{@{} l l l}
\textbf{Scale} & \textbf{Task} & \textbf{Tool} \\
\midrule
Molecular & Single molecule detection & Polaris~\cite{laubscher2024accurate} \\ 
Cellular & Cell segmentation & Cellpose~\cite{stringer2025cellpose3} \\ 
Macroscopic & Medical segmentation & MedSAM~\cite{medsamNatCom} \\ 
\end{tabular}
\end{center}

For each case study, the workflow is built around a provided production-quality domain-specific tool. To enable adaptation and optimization, we include a small labeled validation dataset and a score function to quantify performance. The baselines are established using expert-engineered preprocessing and postprocessing function pairs from the official production-optimized implementations from the original tool authors, requiring weeks to months of tuning effort (quantified via Git history and code complexity analysis in the appendix).


\noindent\textbf{Polaris: Single Molecule Spot Detection.}
This task focuses on detecting sub-pixel fluorescent spots for image-based spatial transcriptomics data, from various modalities using different RNA capturing and tagging methods. We use Polaris \cite{laubscher2024accurate}, a pretrained model for detecting the spots. The expert baseline preprocesses the images with intensity normalization and clipping, and applies peak finding and subpixel localization as postprocessing. The optimization objective is to maximize the F1 score on a validation dataset of 95 images.

\noindent\textbf{Cellpose: Cell Segmentation.}
The objective of this task is cell instance segmentation on multiple modalities such as whole cell \& nucleus, fluorescent and phase bacterial images. We use the ``cyto-3'' model from  Cellpose3 \cite{stringer2025cellpose3}, a U-Net based network pre-trained for general cell segmentation. The expert baseline applies per-channel percentile-based min-max normalization as preprocessing and postprocesses the predictions by removing small objects and filling small holes. The optimization objective is to maximize the average precision at an Intersection over Union (IoU) threshold of 0.5 on a validation dataset of 100 images.

\noindent\textbf{MedSAM: Medical Segmentation.}
This task involves medical image segmentation. MedSAM \cite{medsamNatCom} is an extension of the Segment Anything Model \citep{kirillov2023seganysegment} (SAM) specifically adapted to medical imaging domains. We used a publicly released validation subset from~\citep{medsam-codabench}, focusing on the Dermoscopy modality. The expert functions use per-channel percentile-based scaling for preprocessing and apply bilinear interpolation as postprocessing. The optimization objective is to maximize the sum of the Normalized Surface Dice (NSD) and Dice Similarity Coefficient (DSC) scores on a validation dataset of 25 images.
\section{Results}
\label{sec:results}

\begin{figure*}[h!]
  \centering
  \includegraphics[width=0.95\linewidth]{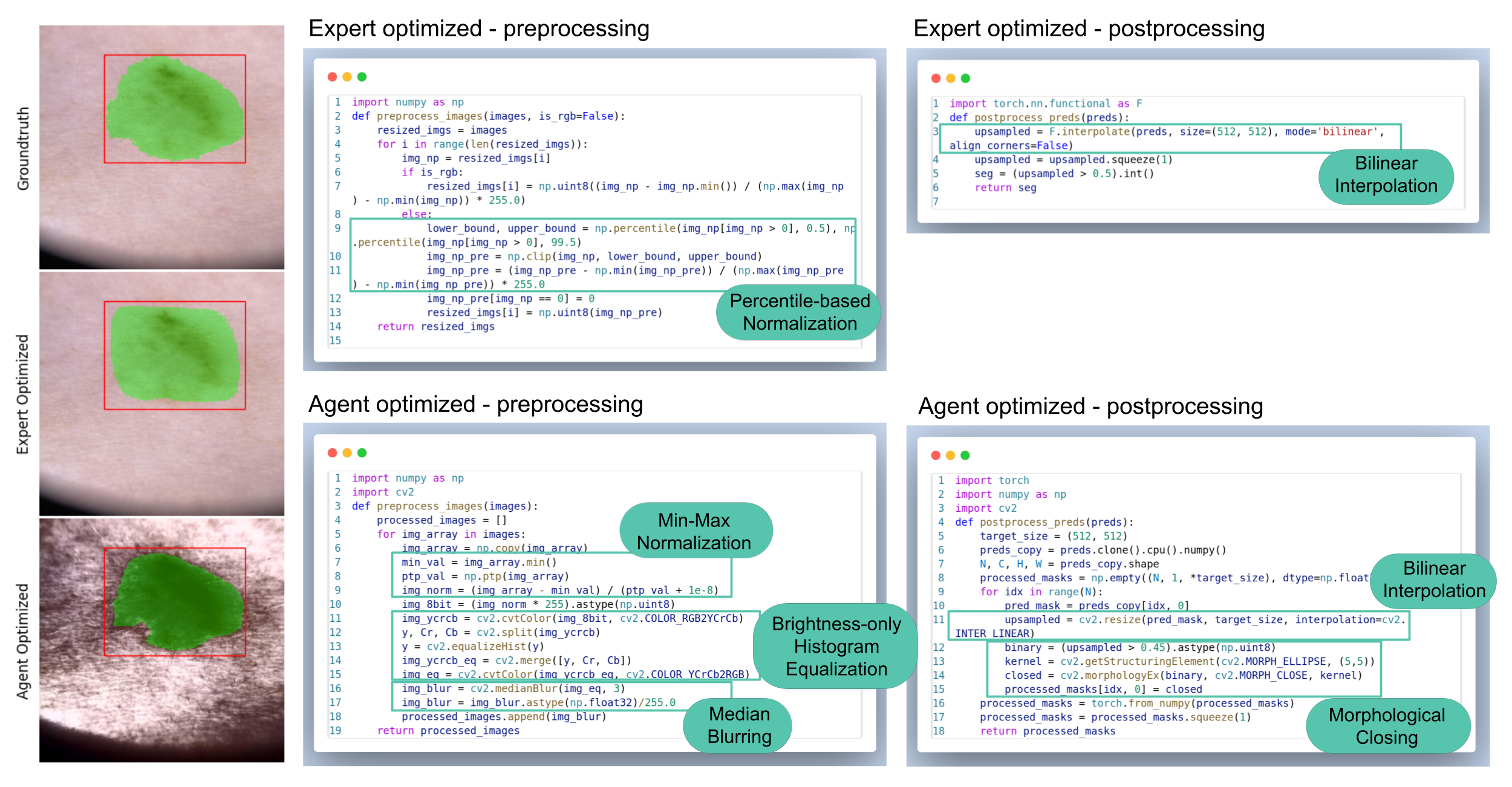}
  \vspace{-0.15in}
  \caption{Comparison of expert-optimized and agent-optimized MedSAM segmentation results. \textit{Left Column}: visual results, showing \textit{(Top)} the raw image with the groundtruth mask and prompt (red box), \textit{(Middle)} the segmentation using expert-optimized functions, and \textit{(Bottom)} the segmentation using agent-generated functions. \textit{Middle \& Right columns}: Comparison of expert \textit{(Top)} and agent-generated \textit{(Bottom)} code for preprocessing and postprocessing, respectively.}
\end{figure*}

\definecolor{good}{HTML}{76ccba} 
\definecolor{bad}{HTML}{f0b56e}  
\definecolor{baseline}{HTML}{faf6d7}  
\newcolumntype{C}[1]{>{\centering\arraybackslash}p{#1}} 

\begin{table*}[h!]
\small
  \centering
  \caption{Design Choice Study. \colorbox{bad}{Orange}: worse than Base Agent, \colorbox{good}{Green}: better than Base Agent. Across all agent settings, all except the ``Small LLM'' outperform ``Expert Baseline''. We observe mixed effects across many design choices.}
\begin{tabular}{l *{8}{c}}
\toprule
& {\makecell[b]{Expert \\ Baseline}}
& {\makecell[b]{Base \\ Agent}}
& {\makecell[b]{Add Expert \\ Function}}
& {\makecell[b]{Add Function \\ Bank}}
& {\makecell[b]{Reasoning \\ LLM}}
& {\makecell[b]{Small \\ LLM}}
& {\makecell[b]{No Data \\ Prompt}}
& {\makecell[b]{No API \\ List}} \\
\midrule
\textbf{Polaris} \scriptsize{(F1)} & \cellcolor{bad} 0.841 & 0.867  & \cellcolor{good} 0.929 & \cellcolor{good} 0.889 & \cellcolor{bad} 0.844 & \cellcolor{bad} 0.805 & \cellcolor{bad} 0.856 & \cellcolor{good} 0.868 \\ 
\textbf{Cellpose} \scriptsize{(AP@IoU 0.5)} & \cellcolor{bad} 0.402 & 0.409  & \cellcolor{good} 0.410 &\cellcolor{good} 0.416 & \cellcolor{good} 0.412 & \cellcolor{bad} 0.397 & \cellcolor{bad} 0.406 & \cellcolor{good} 0.417 \\ 
\textbf{MedSAM} \scriptsize{(NSD+DSC)} & \cellcolor{bad} 0.820 &  0.971 & \cellcolor{bad} 0.888 & \cellcolor{bad} 0.943 & \cellcolor{good} 1.020 & \cellcolor{bad} 0.918 & \cellcolor{bad} 0.952 & \cellcolor{good} 1.037 \\ 
\bottomrule
\end{tabular}
\label{tab:design-choices}
\end{table*}

\subsection{Can simple agents beat expert baselines?}

The main question we ask is: can a simple agent system adapt a fixed, pretrained scientific tool to reliably outperform expert solutions? Our results provide a positive answer (Table~\ref{tab:design-choices}). Across all case studies, the Base Agent is able to generate functions that outperform the expert baselines, with the largest gain seen in MedSAM segmentation. Across all the design choices tested, only the Small LLM (Llama 3.3-70B-Instruct-Turbo) resulted in worse performance compared to the expert baseline. Simple agents provide a practical solution to tool adaptation, finding a solution within 1-2 days of compute time (see appendix for computational details) and saving weeks or months of scientists' manual tuning.

\subsection{How do design choices matter?}

A systematic analysis of agent design choices reveals a striking pattern: most components show inconsistent effects across tasks (Table~\ref{tab:design-choices}). Expert functions dramatically improve Polaris but harm MedSAM. Reasoning LLMs help MedSAM but hurt Polaris. Function Banks, while encouraging diversity, lead to a counterintuitive decline on MedSAM. This inconsistency, set against the prevailing trend of building increasingly complex agents, prompts a re-evaluation of whether architectural sophistication universally yields task-agnostic performance gains.

To understand why a component might benefit one task but harm another, we first introduce a framework for characterizing each problem's solution space along two dimensions: 1) \textbf{API Space}: concentrated (relying on a few commonly co-occurring key APIs) or dispersed (allowing a wider variety of API compositions), and 2) \textbf{Parameter Space}, which can be easy-to-optimize (within the LLM's default bias) or hard-to-optimize (requiring highly specific values).

\subsubsection{Characterizing the Solution Spaces}

\begin{figure*}[h]
  \centering
  \includegraphics[width=0.95\linewidth]{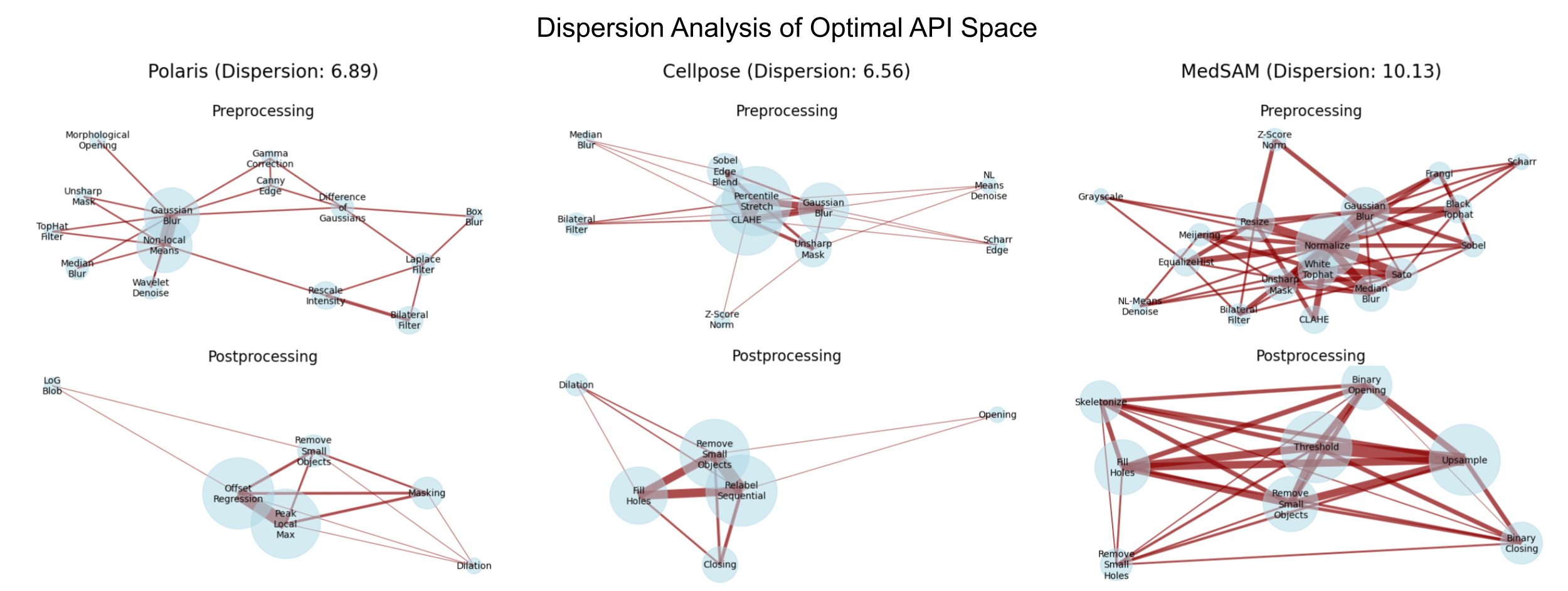}
  \vspace{-0.12in}
  \caption{Optimal API Space Characterization. We analyzed the top 20 functions from all settings, visualizing API call frequency (node size) and co-occurrence frequency (edge weight). The solution spaces for Polaris and Cellpose are highly concentrated, whereas the MedSAM space is significantly more dispersed, with high co-occurrence ratios distributed across the graph. This observation is quantitatively validated by the dispersion score (edge weight entropy), which is notably higher for MedSAM.}
  \label{fig:api-space}
\end{figure*}

\begin{figure*}[h]
  \centering
  \includegraphics[width=0.95\linewidth]{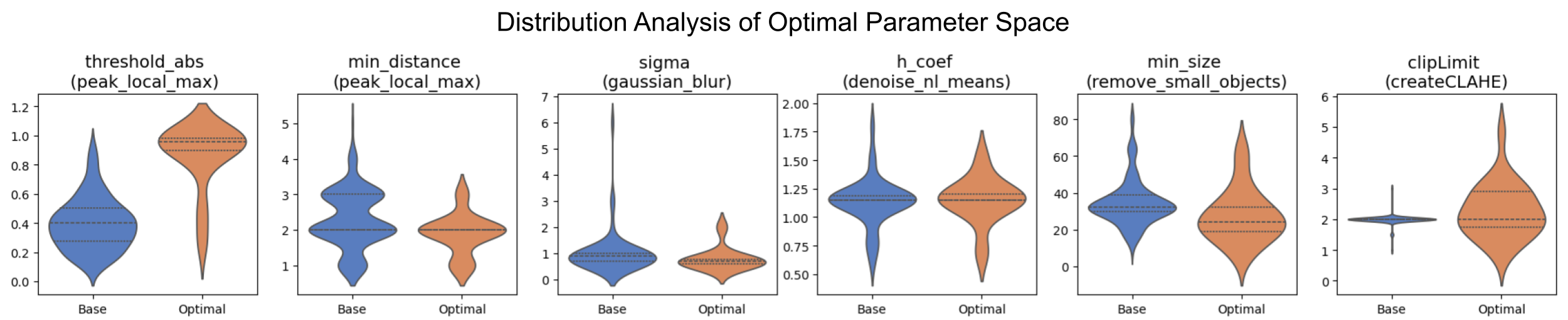}
  \caption{Optimal Parameter Space Characterization. We compared the distributions of six commonly used parameters between the Base Agent setting and found an ``optimal'' range from the top 20 functions across all settings. The distributions were generally similar, with the notable exception of \texttt{\detokenize{threshold_abs}} in \texttt{\detokenize{peak_local_max}}, a commonly used postprocessing parameter in Polaris.}
  \label{fig:param-space}
  \vspace{-0.05in}
\end{figure*}

To map our tasks, we analyzed the top 20 solutions (based on test scores) across all settings to define the ``\textbf{optimal}'' API space and parameter space.

The \textbf{API space} analysis (Fig.~\ref{fig:api-space}) reveals two distinct categories. Polaris and Cellpose have \textbf{concentrated} spaces, dominated by a few strongly connected key APIs, while other APIs are occasionally used. In contrast, MedSAM has a highly \textbf{dispersed} space---with more evenly spread-out strong edges across the graph. The difference is also quantified and validated by the dispersion scores (edge weight entropy, see appendix for the formal definition).

The \textbf{parameter space} analysis (Fig.~\ref{fig:param-space}) shows that the Polaris task is defined by a \textbf{hard-to-optimize} parameter space. This is specifically due to the \verb|threshold_abs| parameter in \verb|peak_local_max|, where we found a drastic, systematic gap between the agent's suggested values and the optimal range. The critical impact of this single parameter was further verified: manually correcting it led to a drastic score improvement (Fig.~\ref{fig:analysis-param}a), confirming the LLM's bias prevented it from finding this highly specified value. In contrast, all other tasks and functions had \textbf{easy-to-optimize} parameter spaces, with agent-proposed distributions aligning well with the optimal ones.

This allows us to categorize our tasks as follows:
\begin{center}
\begin{tabular}{@{} l l l}
 & \textbf{API Space} & \textbf{Parameter Space} \\
\midrule
\textbf{Polaris} & Concentrated & Hard-to-optimize \\ 
\textbf{Cellpose} & Concentrated & Easy-to-optimize \\ 
\textbf{MedSAM} & Dispersed & Easy-to-optimize \\ 
\end{tabular}
\end{center}

\subsubsection{Evaluating Mixed Effects in Context}

\begin{figure}
  \centering
  \includegraphics[width=0.9\linewidth]{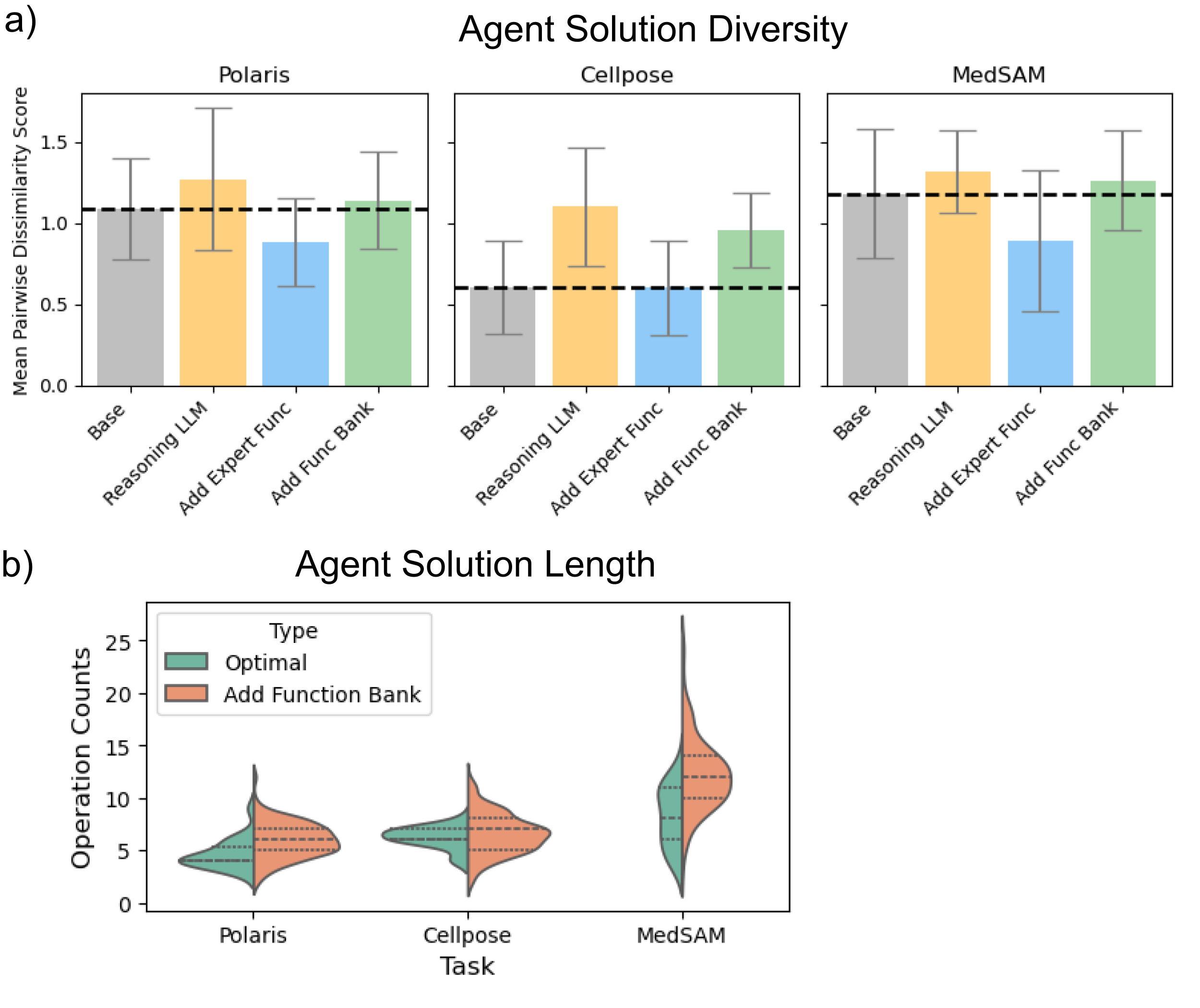}
  \vspace{-0.12in}
  \caption{Analysis of Agent-Generated Solutions. \textit{Top}: Solution diversity, measured by pairwise Jaccard dissimilarity of API sets (see appendix for the formal definition). Both ``Reasoning LLM'' and ``Add Function Bank'' tend to increase diversity whereas ``Add Expert Function'' limits it. \textit{Bottom}: Enabling the Function Bank has a different effect on solution length depending on the task. On concentrated solution spaces (Polaris, Cellpose), there is no obvious effect; however, on the dispersed solution space (MedSAM), the solutions tend to get longer.}
  \label{fig:analysis-api}
\end{figure}

This framework clarifies the mixed results from Table~\ref{tab:design-choices}. 

\noindent\textbf{Expert Functions.} This component is highly beneficial for hard-to-optimize parameter spaces but detrimental to dispersed API spaces. Polaris saw a massive benefit from the added parameter information. Meanwhile, MedSAM (dispersed API) was harmed likely because the component restricted its necessary exploration (Fig.~\ref{fig:analysis-api}a). Cellpose's easy-to-optimize parameter space meant it received no massive boost, while its concentrated API space meant it was not harmed by the restriction, resulting in a moderate positive effect.

\begin{figure}[h]
  \centering
  \includegraphics[width=0.95\linewidth]{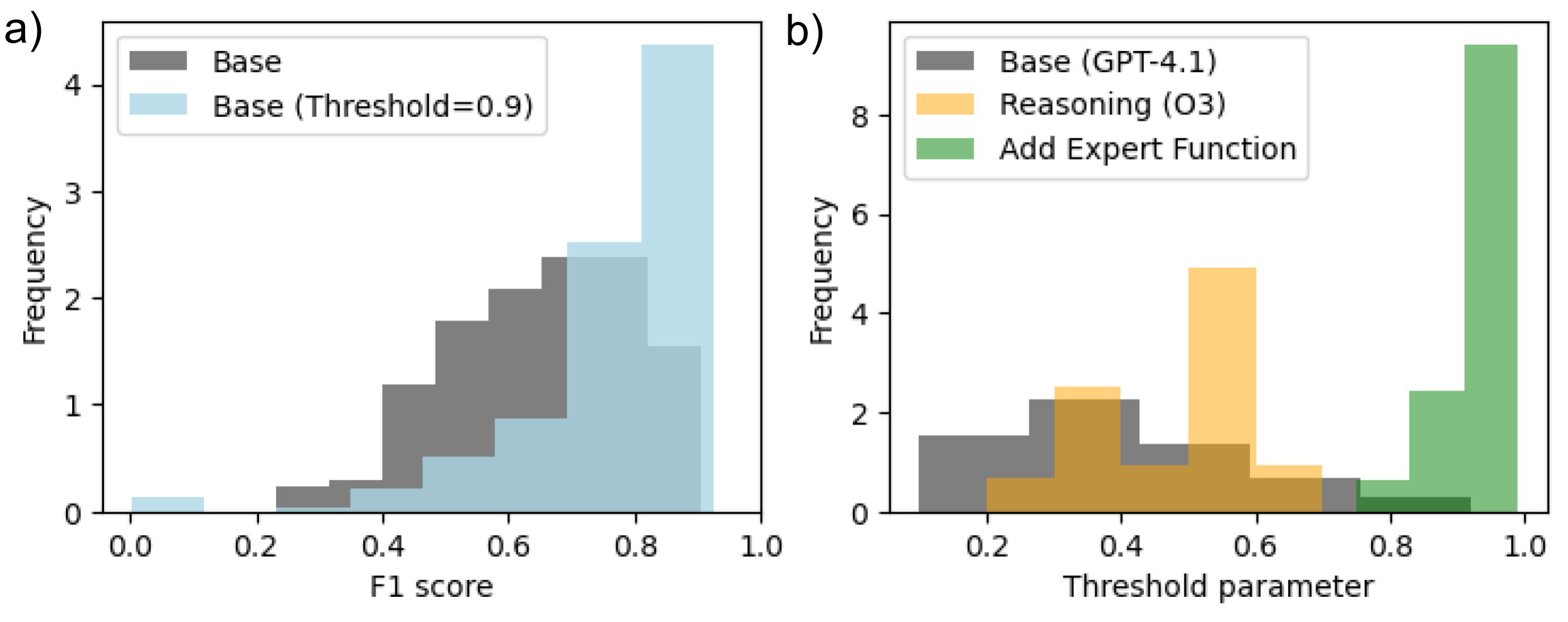}
  \vspace{-0.12in}
  \caption{Detailed Analysis of Polaris \texttt{\detokenize{threshold_abs}} Parameter. \textit{Left}: Manually setting the parameter to 0.9 increased scores. \textit{Right}: Agent analysis shows expert functions facilitate finding the optimal range. Reasoning LLMs suggested a smaller, more centralized parameter range (biased towards 0.5 in 49.32\% of suggestions) than General LLMs (10.43\%).}
  \vspace{-0.05in}
  \label{fig:analysis-param}
\end{figure}

\noindent\textbf{Reasoning LLM.} The Reasoning LLM's impact is best understood by its uneven exploratory behavior: it excelled at function diversity but failed at parameter search. This enhanced diversity (Fig.~\ref{fig:analysis-api}a) is likely beneficial for MedSAM's dispersed API space. Conversely, it is more constrained in its parameter choices (Fig.~\ref{fig:analysis-param}b), preventing it from finding the optimal parameters on Polaris. As before, the more neutral Cellpose task saw a moderate performance boost.

\noindent\textbf{Function Bank.} The Function Bank's diversity-boosting effect (Fig~\ref{fig:analysis-api}a) acted as a double-edged sword. While it led to better performance for Cellpose and Polaris, it surprisingly hurt MedSAM scores. We observed that, in a dispersed space, the component encourages the agent to build progressively longer solutions (Fig~\ref{fig:analysis-api}b), which may eventually become detrimental.

\subsubsection{Stable Design Choices Across All Tasks}

\begin{figure}[h]
  \centering
  \includegraphics[width=0.9\linewidth]{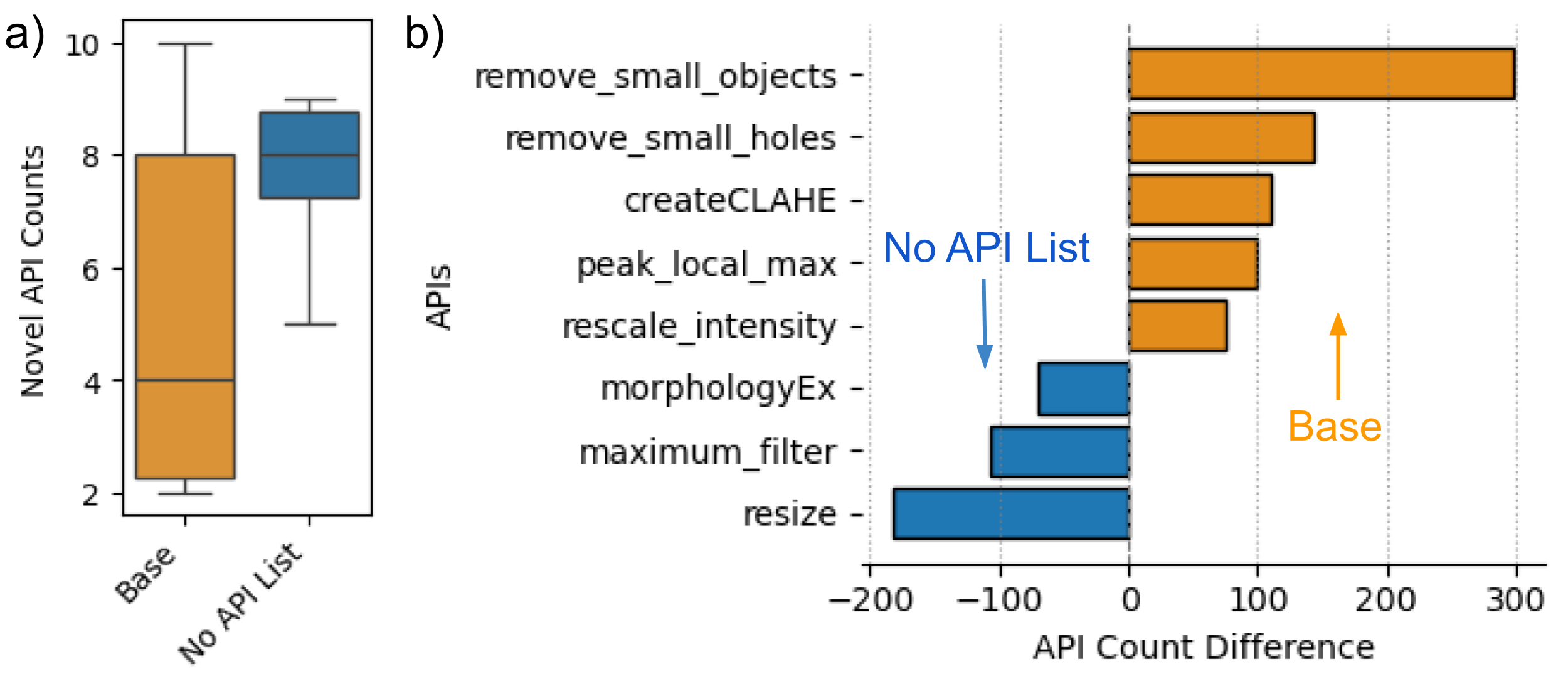}
  \vspace{-0.12in}
  \caption{Effect of API List on Agent Behavior. \textit{Left}: Numbers of novel APIs discovered by the agent with and without the provided API list. \textit{Right}: API usage difference between the two settings.}
  \label{fig:ablate-api-list}
  \vspace{-0.1in}
\end{figure}

We also identified two design choices that produced stable, consistent results across all tasks (Table~\ref{tab:design-choices}). 

\noindent\textbf{Data Prompt.} Ablating the data prompt consistently worsened performance, which suggests that the data context is essential for agents to write suitable functions. 

\noindent\textbf{API List.} Ablating the API list consistently improved scores. Further analysis showed that agents can discover novel APIs regardless of the presence of an API list, but do so more consistently without it (Fig.~\ref{fig:ablate-api-list}a), suggesting that the LLM's latent knowledge might be sufficient for solving these tasks. Additionally, providing the list appears to introduce a harmful bias (Fig.~\ref{fig:ablate-api-list}b), evidenced by the unexpected high usage of specific APIs like \verb|remove_small_objects| and \verb|remove_small_holes|. While these APIs are relevant and are used even when the list is omitted, their usage became strongly and disproportionately biased when the API list was provided. This suggests that the default choice should be to omit the list unless the APIs are beyond the LLM's intrinsic knowledge scope.

\subsection{Can AutoML help search the parameter space more effectively?}

We investigated whether introducing an AutoML agent  component (defined in \ref{sec:approach-advanced-components}) could overcome the LLM's bias in hard-to-optimize parameter spaces. However, our findings show that it is not a straightforward remedy and introduces new, significant challenges, such as overfitting.

First, a simple, non-agentic baseline performed worse on all three tasks (Table~\ref{tab:automl}). For this baseline, we prompted the LLM in a single shot to write a complete AutoML template (details in appendix), which is then executed using the same tool-call budget as the agent studies. This approach proved ineffective; the resulting search was neither comprehensive nor optimal, with the LLM identifying only 4.8 ($\pm$ 1.5) functions to optimize on average.

Second, incorporating AutoML directly into the agent framework yielded mixed results, improving MedSAM scores but worsening them on Polaris. The performance drop on Polaris was caused by overfitting on the validation set. An ablation study (Table~\ref{tab:automl-ablate}) confirmed this hypothesis: when we decreased the frequency of AutoML runs or reduced the number of trials, the validation scores dropped, but the final test scores increased. This indicates that excessive hyperparameter optimization is counterproductive, highlighting the need to develop more balanced AutoML search protocols.

\begin{table}[h]
\small
  \centering
\caption{AutoML Optimization Study. Non-agentic AutoML resulted in worse performance compared to the Base Agent. When applying an AutoML search to the Function Bank yielded mixed results: it improved MedSAM performance but surprisingly degraded performance on Polaris.}
\begin{tabular}{@{}lccc}
\toprule
& \makecell{Non-Agentic\\ AutoML} & \makecell{Agent w/\\FuncBank} & \makecell{Agent w/ FuncBank\\ + AutoML} \\
\midrule
\textbf{Polaris}  & 0.844 & 0.889 & 0.877 \\ 
\textbf{Cellpose} & 0.373 & 0.416 & 0.417\\
\textbf{MedSAM}  & 0.879 & 0.943 & 1.014 \\ 
\bottomrule
\end{tabular}
\label{tab:automl}
\end{table}

\begin{table}[h]
\vspace{-0.05in}
\small
  \centering
\caption{Ablation Study on AutoML. Reducing AutoML search frequency or number of iterations decreases validation scores but improves test performance. This beneficial effect is most pronounced when the number of search iterations is reduced.}
\vspace{-0.05in}
\begin{tabular}{lccc}
\toprule
& {Full} & {Half Frequency} & {Half Iterations} \\ 
\midrule
\textbf{Val} \scriptsize{(Polaris)} & 0.733 & 0.702 & 0.718 \\ 
\textbf{Test} \scriptsize{(Polaris)} & 0.877 & 0.890 & 0.910 \\ 
\bottomrule
\end{tabular}
\label{tab:automl-ablate}
\vspace{-0.05in}
\end{table}

\subsection{How do minimal agents compare to a complex tree-search agent?}

\begin{table}[h]
\small
  \centering
\caption{Comparison of Minimal Agents with Proprietary Tree-Search-Based AIDE Agent.}
\begin{tabular}{lccc}
\toprule
& {Base Agent} & \makecell{Base Agent \\ w/ FuncBank} & \makecell{AIDE Agent \\ (Tree Search)} \\ 
\midrule
\textbf{Polaris} & 0.867 & 0.889 & 0.872 \\ 
\textbf{Cellpose}  & 0.409 & 0.416 & 0.414 \\ 
\textbf{MedSAM} & 0.971 & 0.943 & 0.971 \\
\bottomrule
\end{tabular}
\label{tab:aide}
\vspace{-0.05in}
\end{table}

To determine if more complex agent frameworks provide an advantage, we benchmarked the minimal agents against the proprietary AIDE agent~\cite{jiang2025aide}. Conceptually, AIDE's tree search can be viewed as a more sophisticated extension of the Function Bank component, as it integrates prior functions within a structured search space to guide exploration.

To ensure a fair comparison, we calibrated the experimental budget based on the count of valid solutions. Our agents' 60 trials produced 60 valid solutions. For AIDE, as its trials include buggy nodes, we configured it to run 80 trials, which yielded comparable numbers of valid solutions across tasks (MedSAM: 60.9 $\pm$ 5.3, Cellpose: 69.9 $\pm$ 6.7, Polaris: 70.7 $\pm$ 3.3).

Despite its advanced search policy, AIDE exhibited no significant performance advantage, compared with the two minimal agent settings (Table~\ref{tab:aide}). Overall, these findings indicate that for the tool adaptation task, the additional complexity of tree search does not confer a clear out-of-the-box advantage. A minimal, open-source agent framework may achieve comparable results with greater transparency and cost efficiency, offering a more accessible and practical starting point for tool adaptation in real-life scientific applications.

\section{Discussion}
\label{sec:conclusion}

\begin{figure}[t!]
  \centering
  \includegraphics[width=1.0\linewidth]{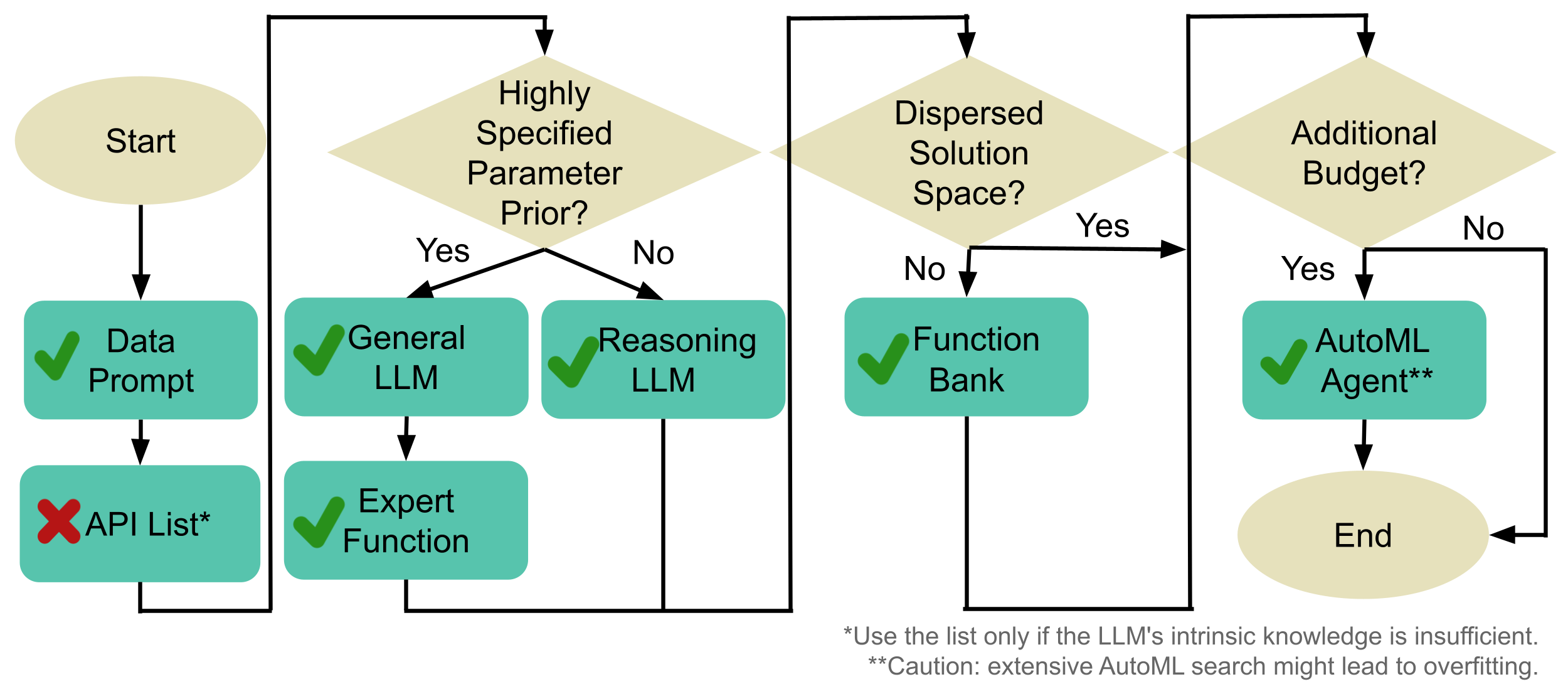}
  \caption{Roadmap on Agent Designs for Tool Adaptation.}
  \label{fig:flowchart}
\end{figure}

\textbf{Roadmap on Tool Adaptation Agent Designs.} We distill our findings into a  recommendations flowchart (Fig.~\ref{fig:flowchart}) as a roadmap for building tool adaptation agents. It illustrates the paths supported by our findings; configurations not depicted are outside the scope of our recommendations based on the current results.\footnote{As foundation models and agent design continue to improve, this roadmap will also evolve over time. The systematic evaluation framework introduced in this work provides a methodology for updating these recommendations as the underlying technologies advance.}

\noindent\textbf{Limitations and Future Work.} First, we focus exclusively on biomedical imaging applications—broadening to other scientific domains would test whether our findings generalize beyond this field. Furthermore, our analysis identified a critical challenge for this low-data regime: the agent's tendency to overfit small validation sets. This finding encourages future research to develop agents that are more resilient to noisy reward signals. Finally, our study intentionally focused on simple, minimal agent designs to establish a rigorous baseline. This provides a controlled foundation for future work to systematically explore if, and when, more complex designs (e.g., RAG, multi-step workflows, or multi-agent systems) can provide justifiable performance gains.

\noindent\textbf{Conclusion.} We investigated the ``last mile'' adaptation bottleneck in biomedical imaging tasks and demonstrated that a minimal ``Base Agent'' consistently outperforms expert-engineered solutions, reducing adaptation time from weeks or months to just 1-2 days. Our systematic analysis reveals that added agent complexity is not universally beneficial; instead, the optimal agent design is context-dependent. Ultimately, this work provides a practical roadmap and validated open-source framework, culminating in a minimal agent solution successfully merged into production, proving its effectiveness as a starting point for real-world tool adaptation.

\section*{Acknowledgments}

For JJS, this study was supported by NSF Award IIS-2505098 and the Food and Drug Administration (FDA) of the U.S. Department of Health and Human Services (HHS) as part of a financial assistance award (U01FD008421) totaling \$199,907 with $100\%$ percent funded by FDA/HHS. The contents are those of the authors and do not necessarily represent the official views of, nor an endorsement, by FDA/HHS, or the U.S. Government. YY was supported by NSF Award 2505096 and in part by a gift from OpenAI. We also would like to thank Kristin Branson for her support.

{
    \small
    \bibliographystyle{ieeenat_fullname}
    \bibliography{main}
}

\clearpage
\setcounter{page}{1}
\appendix
\maketitlesupplementary

The sections of our appendix are organized as follows:

\begin{itemize}
    \item Section \ref{supp:visualizations}: Qualitative results (visualization of the expert and agent processed images and results).
    \item Section \ref{supp:deployment}: Agent function deployed to production (Anonymous GitHub PR screenshot).
    \item Section \ref{supp:exprt-and-agent-funtions}: Expert functions and agent generated functions.
    \item Section \ref{supp:git_history} Git History Analysis of expert baseline functions.
    \item Section \ref{supp:analysis-metrics}: Analysis metric details.
    \item Section \ref{supp:dataset}: Additional dataset details.
    \item Section \ref{supp:prompt}: Prompt details.
    \item Section \ref{supp:single-shot-automl}: Non-Agentic AutoML Baseline.
    \item Section \ref{supp:aide}: AIDE Baseline.
    \item Section \ref{supp:comp-requirement}: Computational requirements.
    \item Section \ref{supp:api-list}: API list.
\end{itemize}

\section{Visualizations}
\label{supp:visualizations}
\subsection{Polaris}

\begin{figure}[H]
    \centering
    \vspace{-0.2in}
    \includegraphics[width=0.9\linewidth]{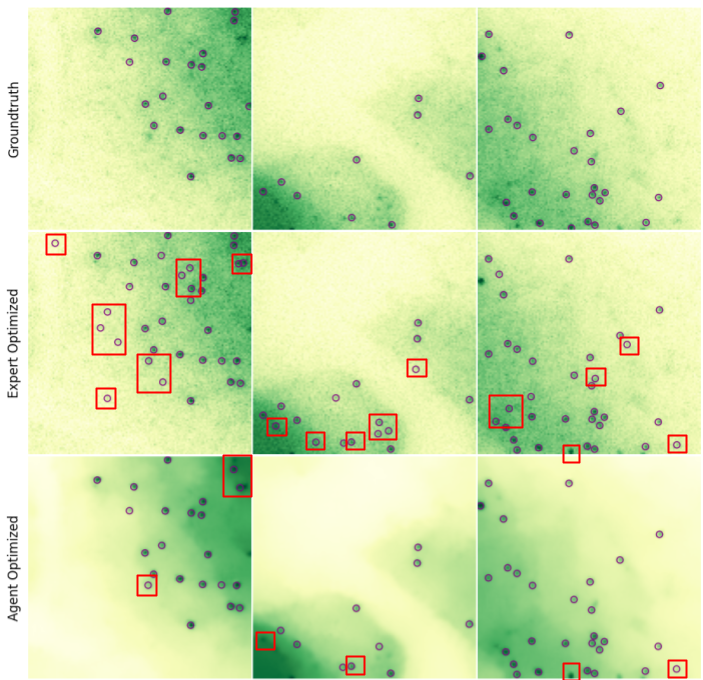}
    \caption{Example images showing (Top) Groundtruth, (Middle) Predictions using expert-optimized function pairs, and (Bottom) Prediction using agent-optimized function pairs. Prediction errors are marked with red boxes.}
    \vspace{-0.2in}
\end{figure}

\subsection{Cellpose}

\begin{figure}[H]
    \centering
    \vspace{-0.2in}
    \includegraphics[width=0.9\linewidth]{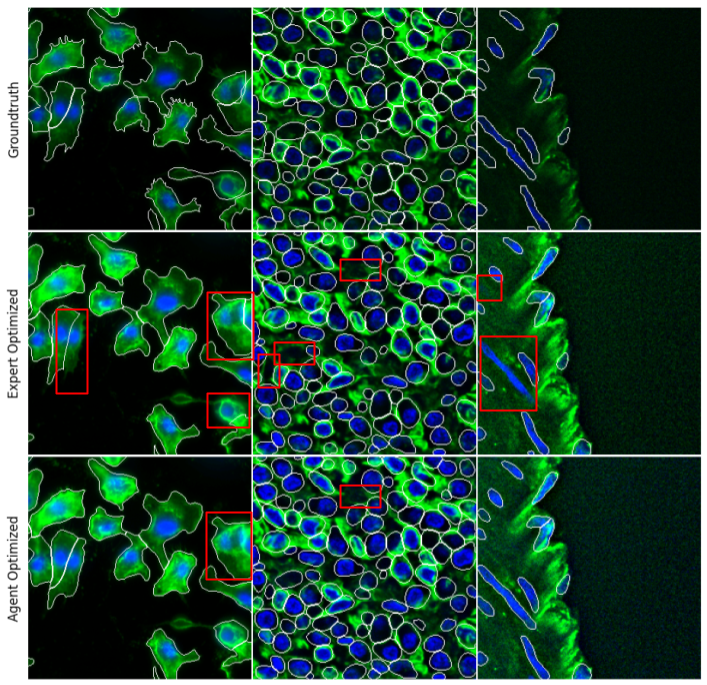}
    \caption{Example images showing (Top) Groundtruth, (Middle) Predictions using expert-optimized function pairs, and (Bottom) Prediction using agent-optimized function pairs. Prediction errors are marked with red boxes.}
    \vspace{-0.2in}
\end{figure}

\subsection{MedSAM}

\begin{figure}[H]
    \centering
    \vspace{-0.2in}
    \includegraphics[width=0.9\linewidth]{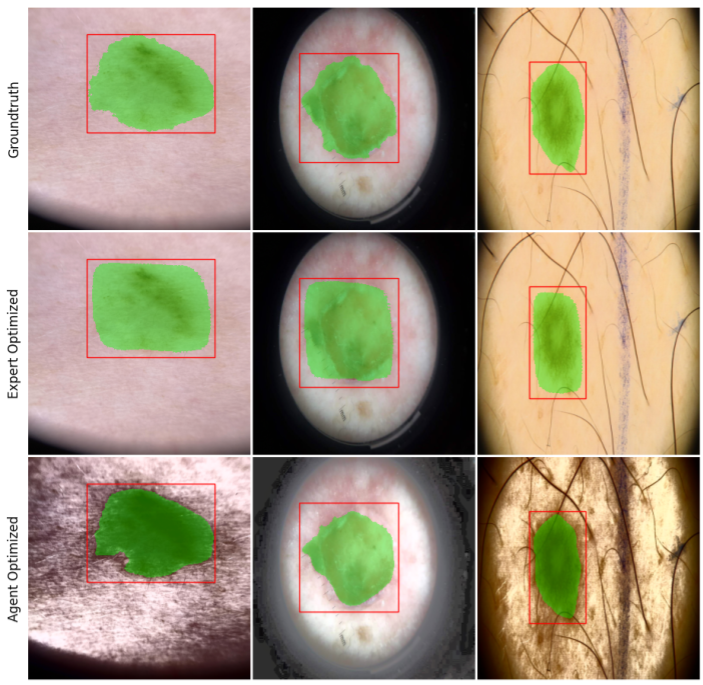}
    \caption{Example images showing (Top) Prompt box and groundtruth, (Middle) Prompt box and prediction using expert-optimized function pairs, and (Bottom) Prompt box and prediction using agent-optimized function pairs. Prediction errors are marked with red boxes.}
    \vspace{-0.2in}
\end{figure}

\section{Agent function deployed to production}
\label{supp:deployment}

\begin{figure}[H]
    \centering
    \includegraphics[width=1.0\linewidth]{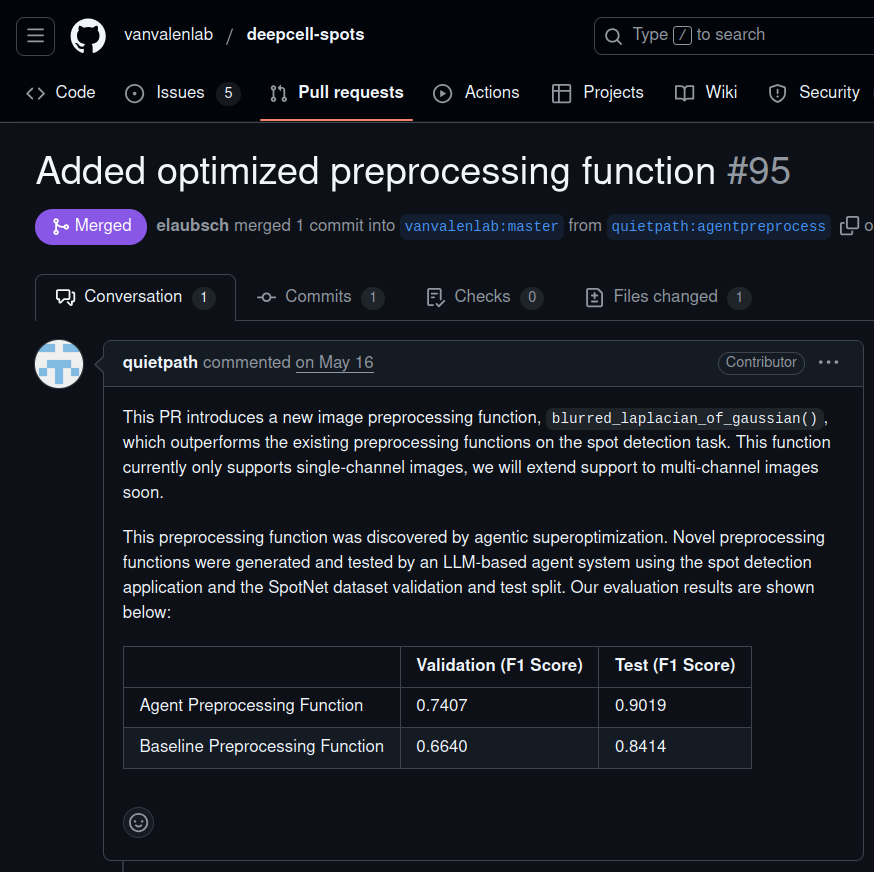}
    \caption{Anonymous authored GitHub PR screenshot of the agent generated functions being integrated into the official codebase.}
\end{figure}

\section{Expert and agent generated functions}
\label{supp:exprt-and-agent-funtions}

\subsection{Polaris}

\begin{figure}[H]
    \centering
    \includegraphics[width=1.0\linewidth]{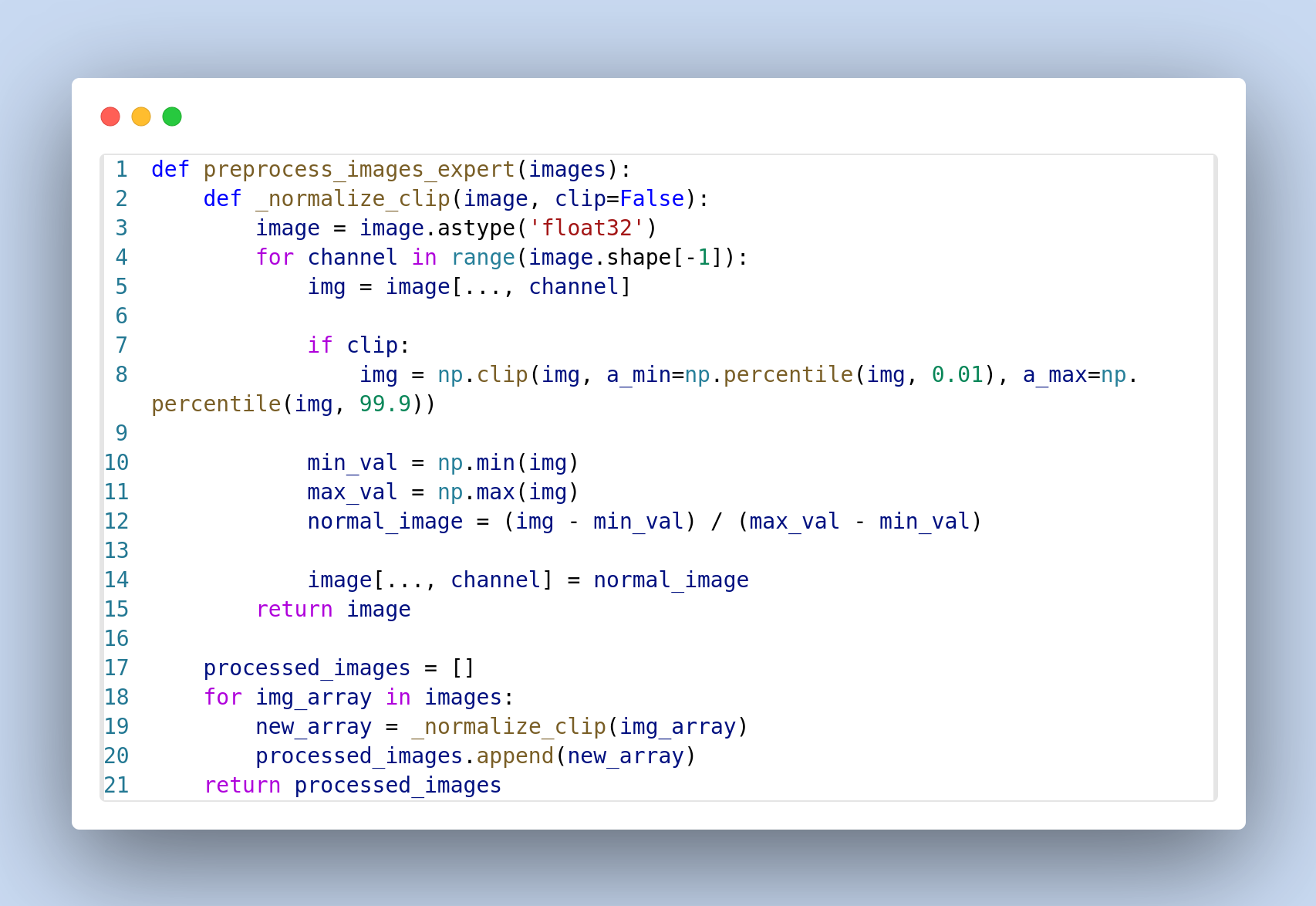}
    \caption{Polaris Expert Preprocessing Function}
\end{figure}

\begin{figure}[H]
    \centering
    \includegraphics[width=1.0\linewidth]{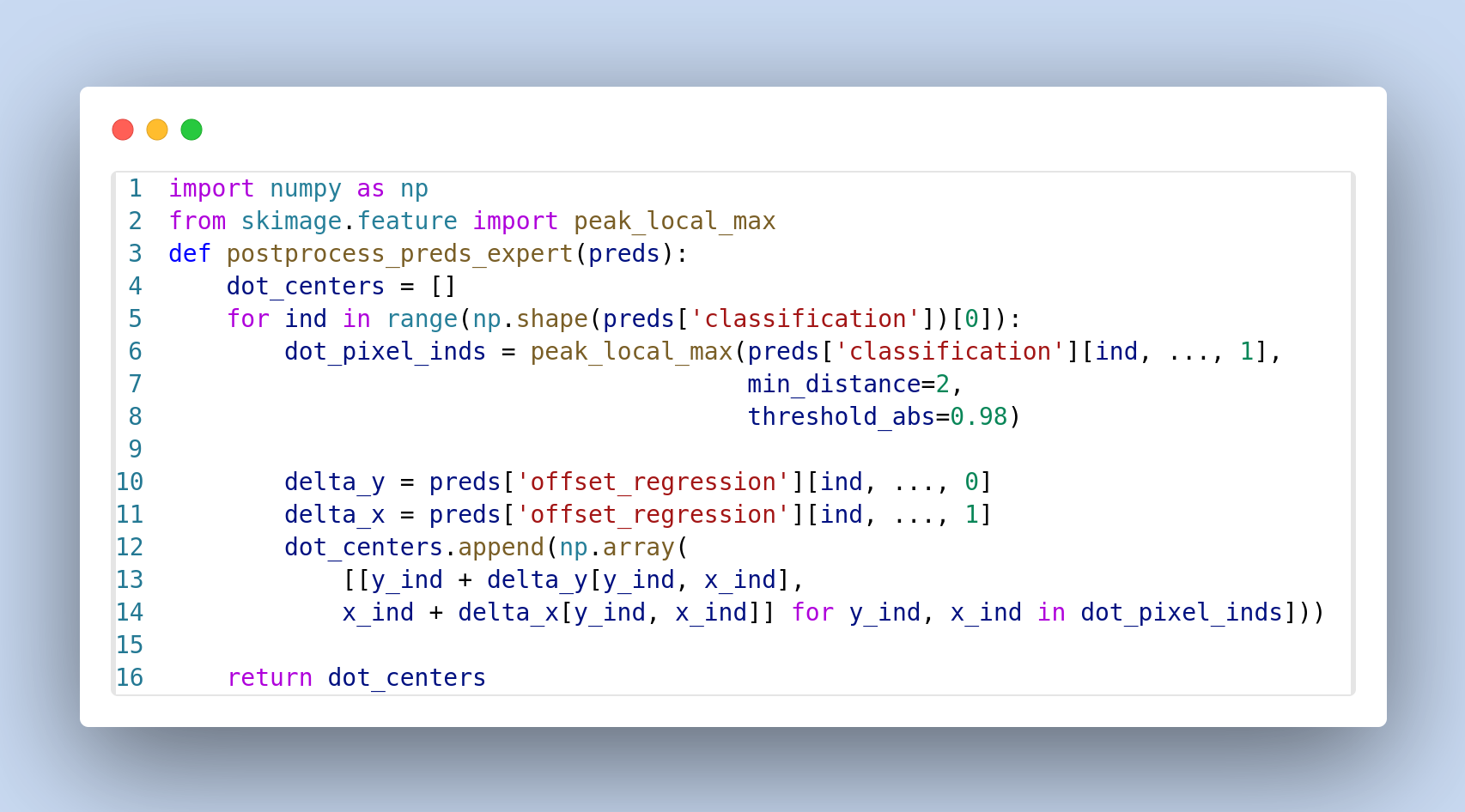}
    \caption{Polaris Expert Postprocessing Function}
\end{figure}

\begin{figure}[H]
    \centering
    \includegraphics[width=1.0\linewidth]{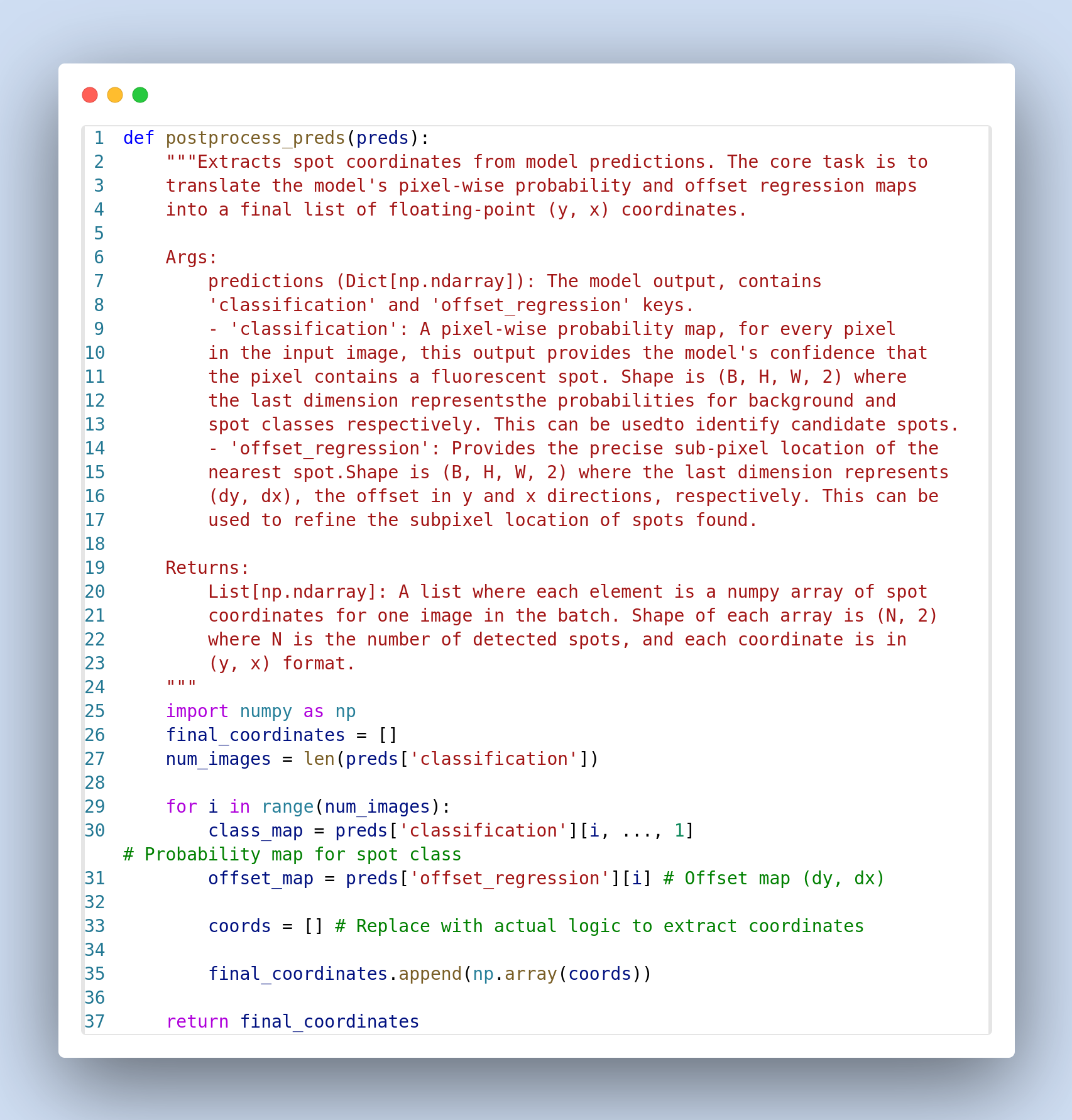}
    \caption{Polaris Postprocessing Function Skeleton}
\end{figure}

\begin{figure}[H]
    \centering
    \includegraphics[width=1.0\linewidth]{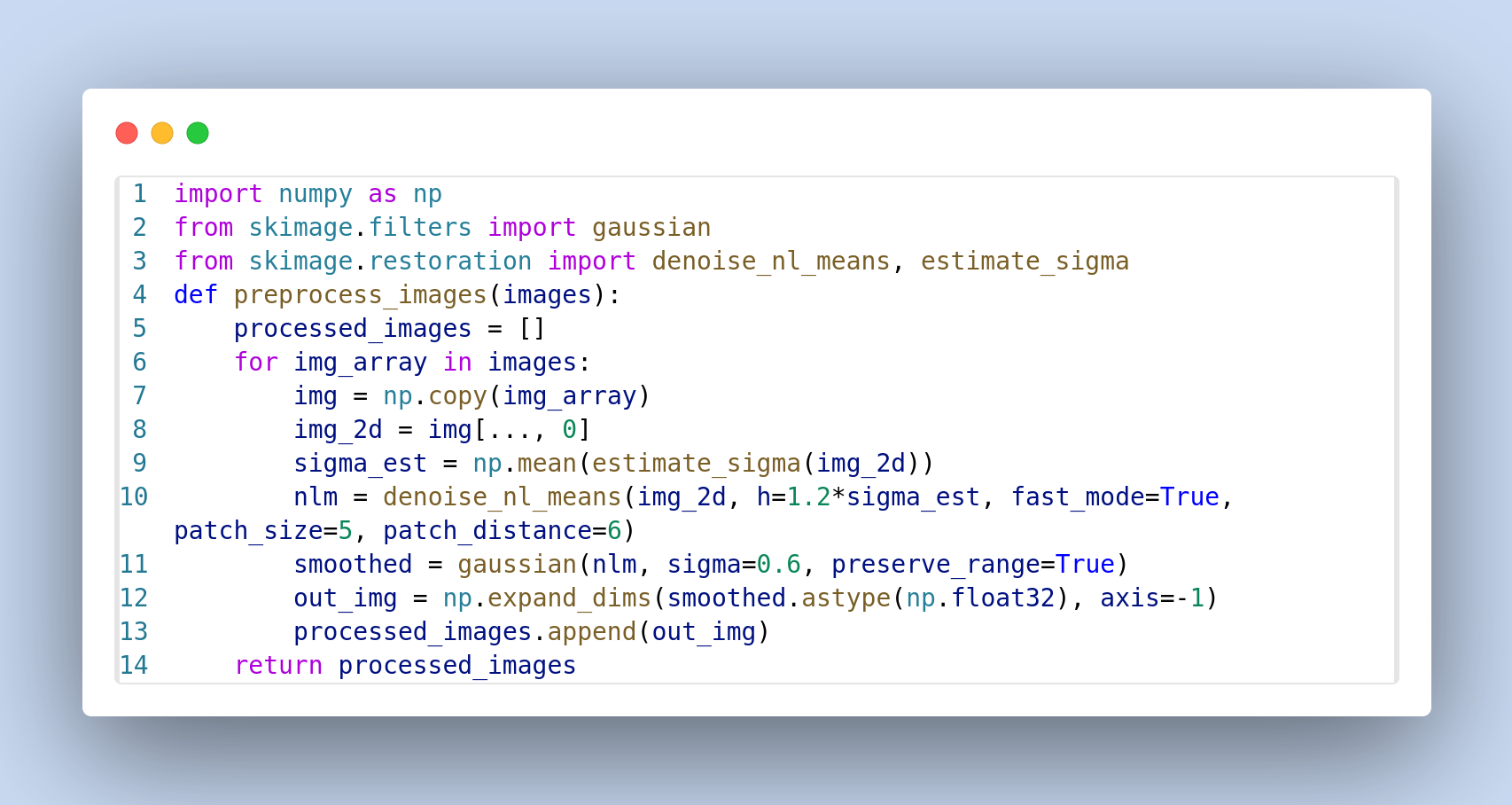}
    \caption{Polaris Agent Generated Preprocessing Function}
\end{figure}

\begin{figure}[H]
    \centering
    \includegraphics[width=1.0\linewidth]{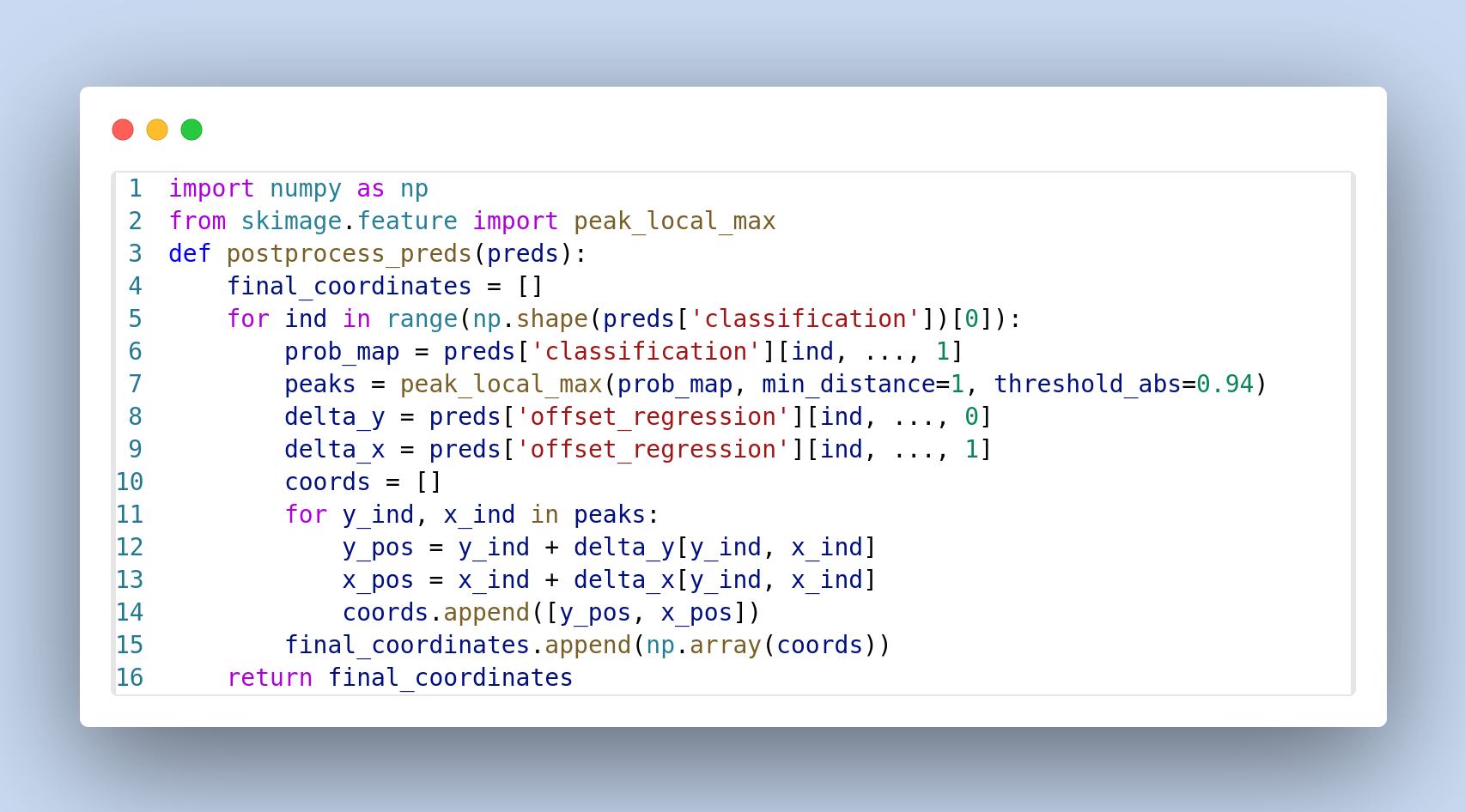}
    \caption{Polaris Agent Generated Postprocessing Function}
\end{figure}

\subsection{Cellpose}

\begin{figure}[H]
    \centering
    \includegraphics[width=1.0\linewidth]{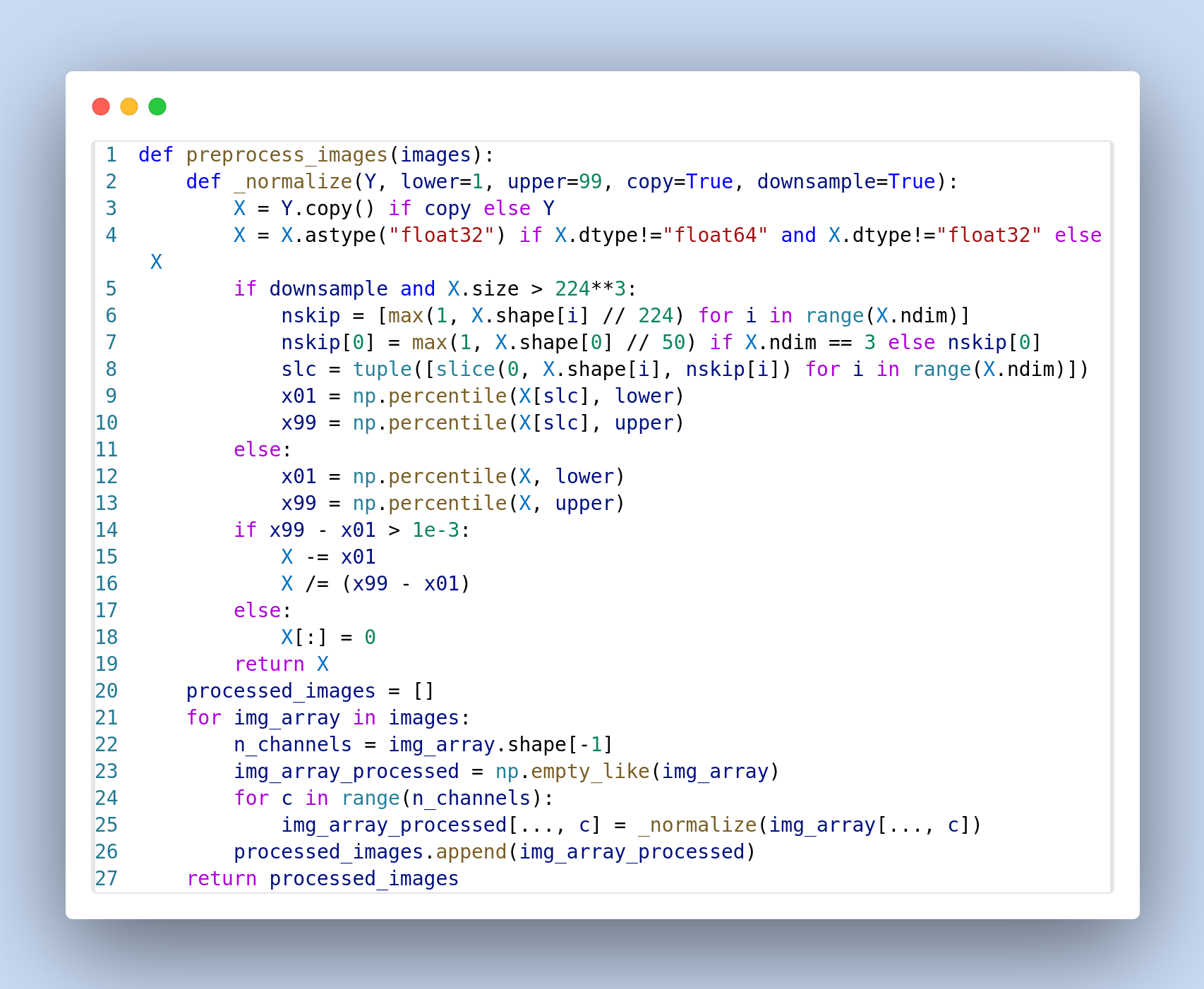}
    \caption{Cellpose Expert Preprocessing Function}
\end{figure}

\begin{figure}[H]
    \centering
    \includegraphics[width=1.0\linewidth]{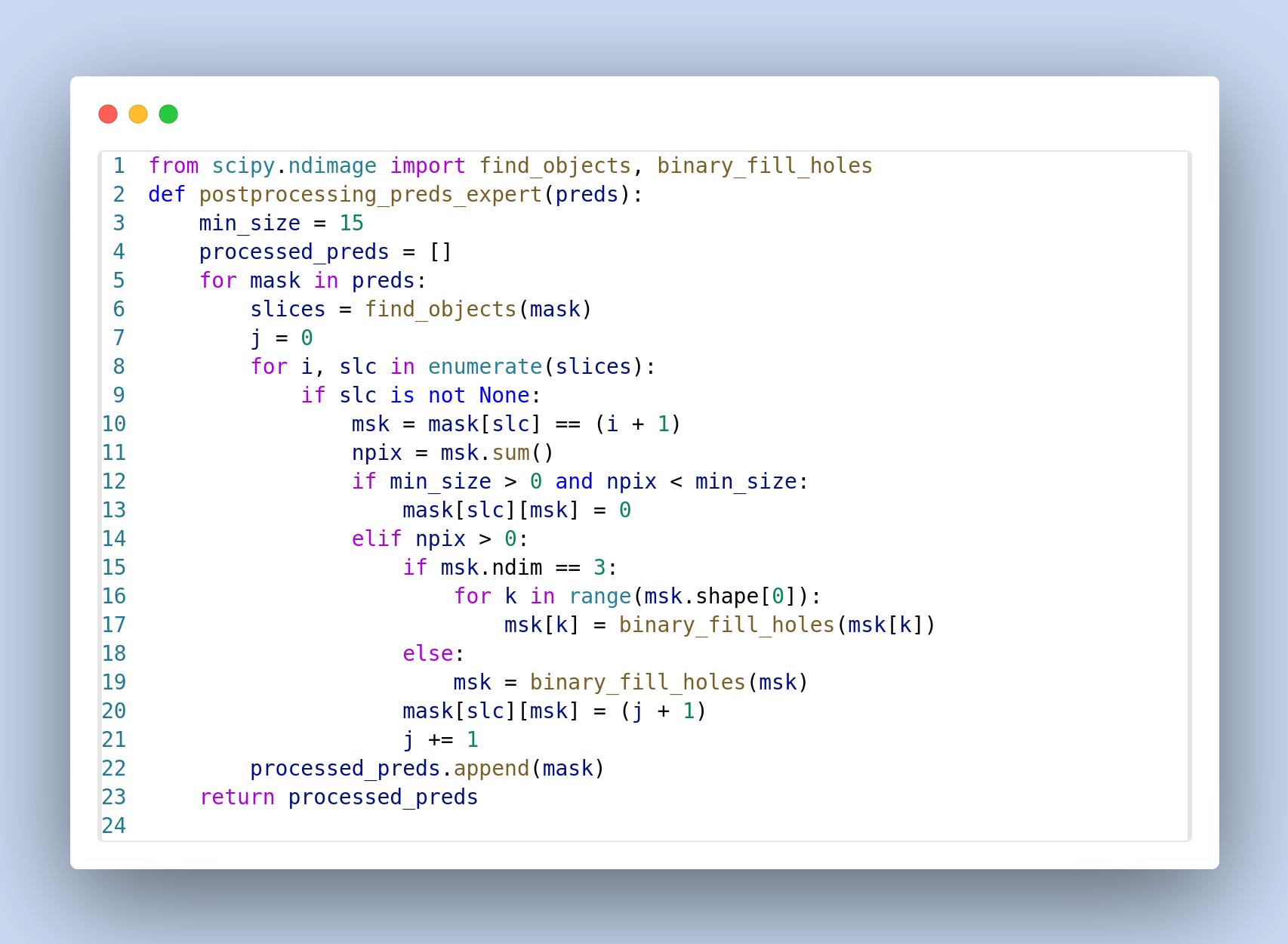}
    \caption{Cellpose Expert Postprocessing Function}
\end{figure}

\begin{figure}[H]
    \centering
    \includegraphics[width=1.0\linewidth]{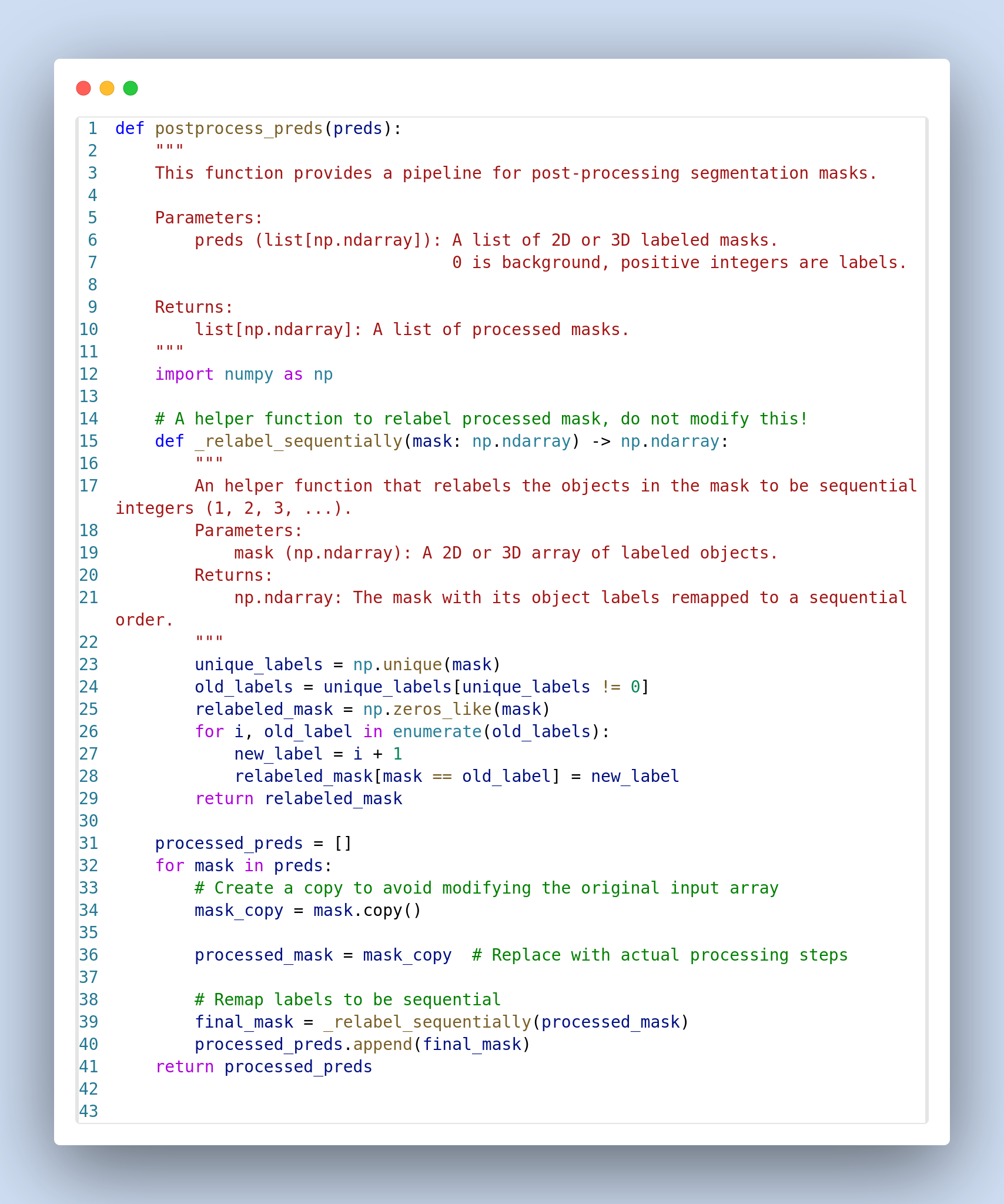}
    \caption{Cellpose Postprocessing Function Skeleton}
\end{figure}

\begin{figure}[H]
    \centering
    \includegraphics[width=1.0\linewidth]{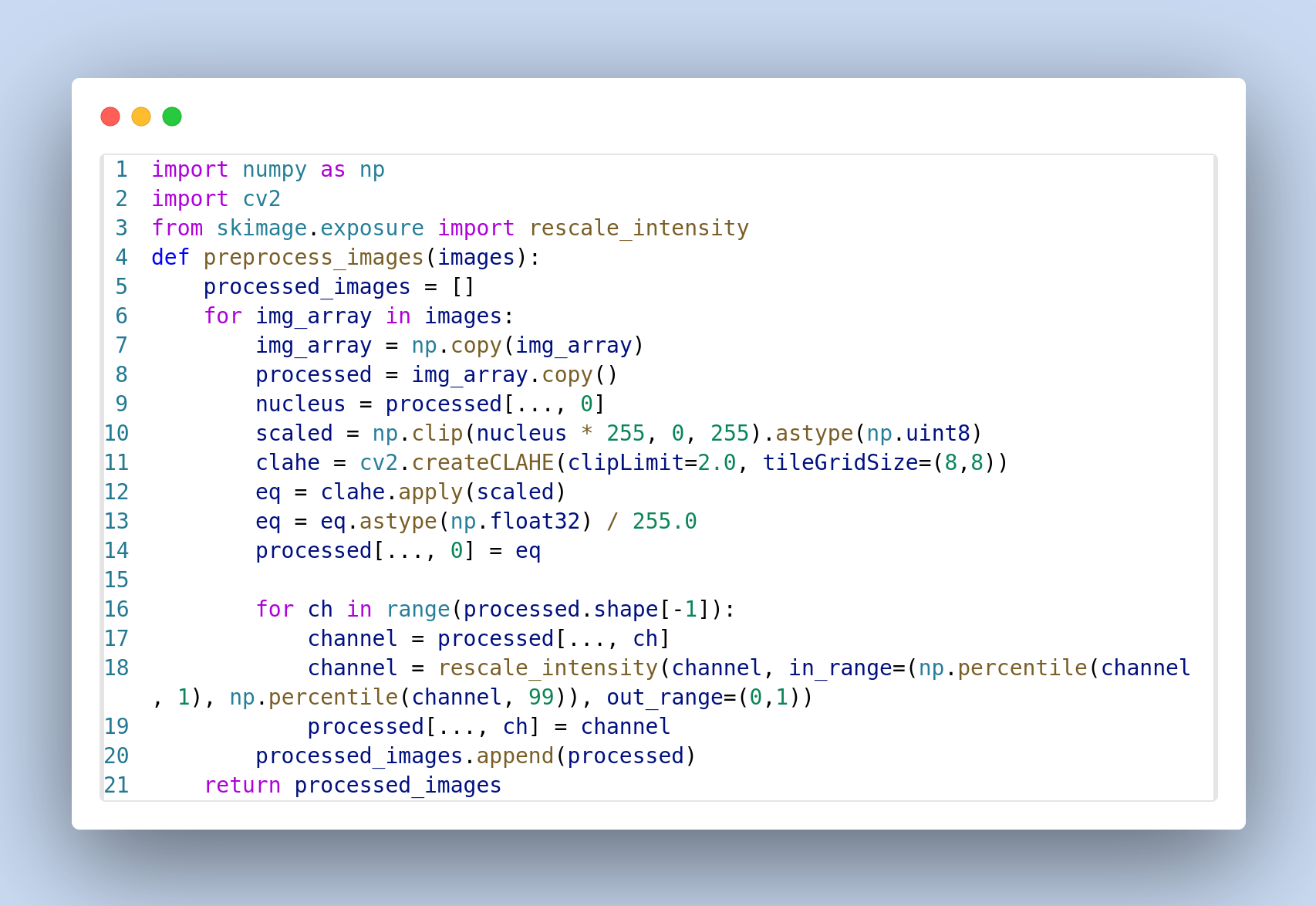}
    \caption{Cellpose Agent Generated Preprocessing Function}
\end{figure}

\begin{figure}[H]
    \centering
    \includegraphics[width=1.0\linewidth]{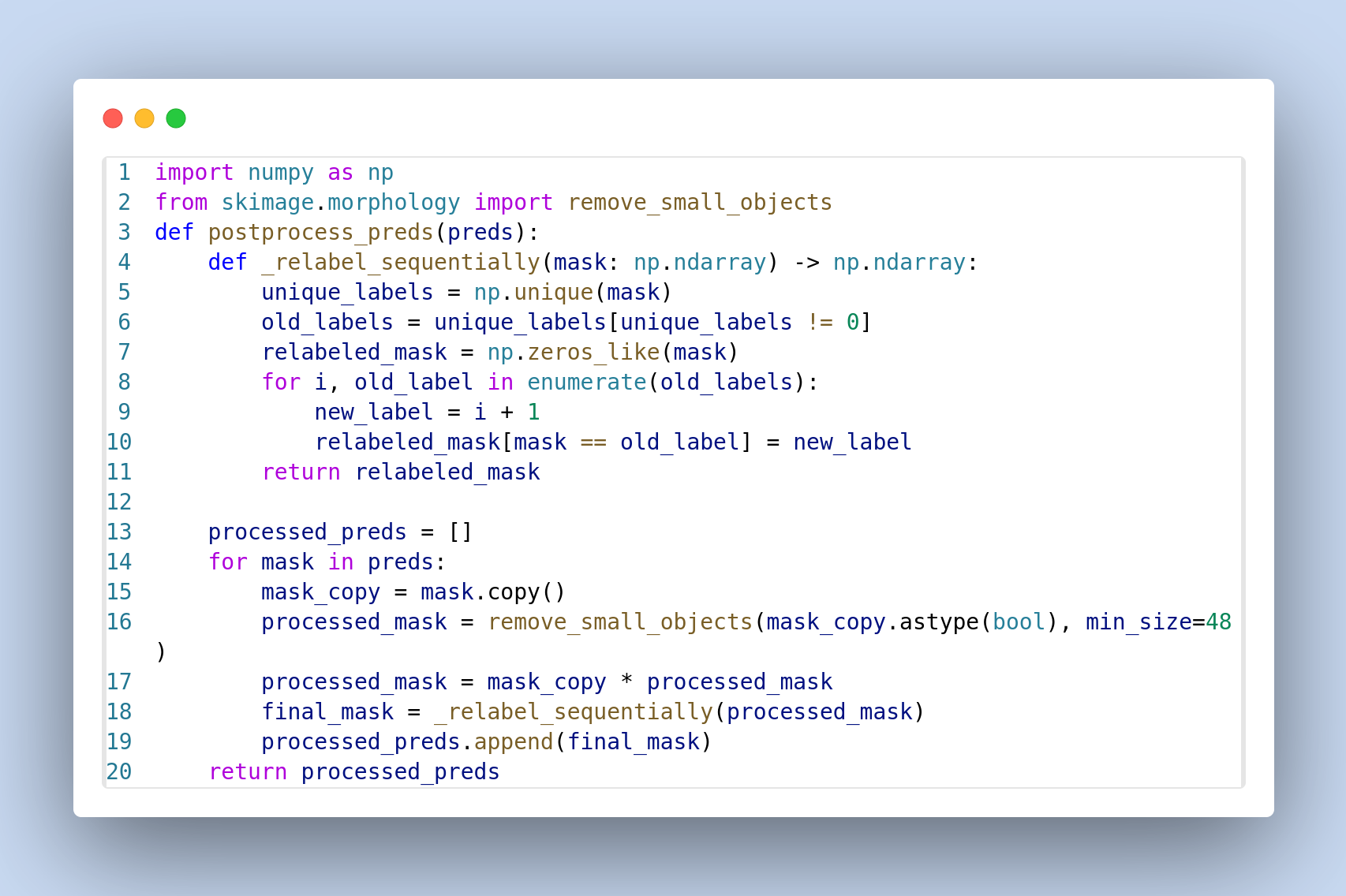}
    \caption{Cellpose Agent Generated Postprocessing Function}
\end{figure}

\subsection{MedSAM}

\begin{figure}[H]
    \centering
    \includegraphics[width=1.0\linewidth]{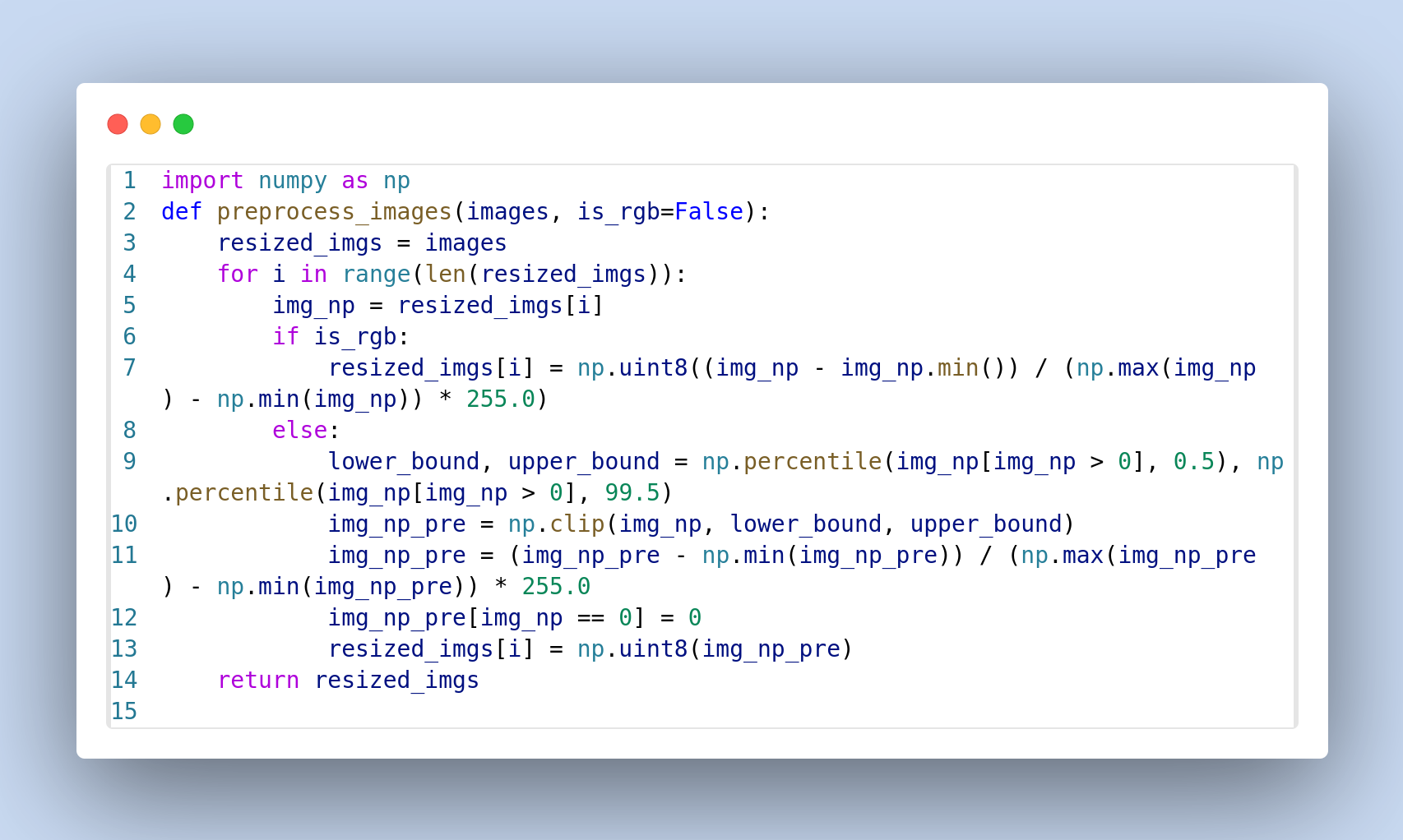}
    \caption{MedSAM Expert Preprocessing Function}
\end{figure}

\begin{figure}[H]
    \centering
    \includegraphics[width=1.0\linewidth]{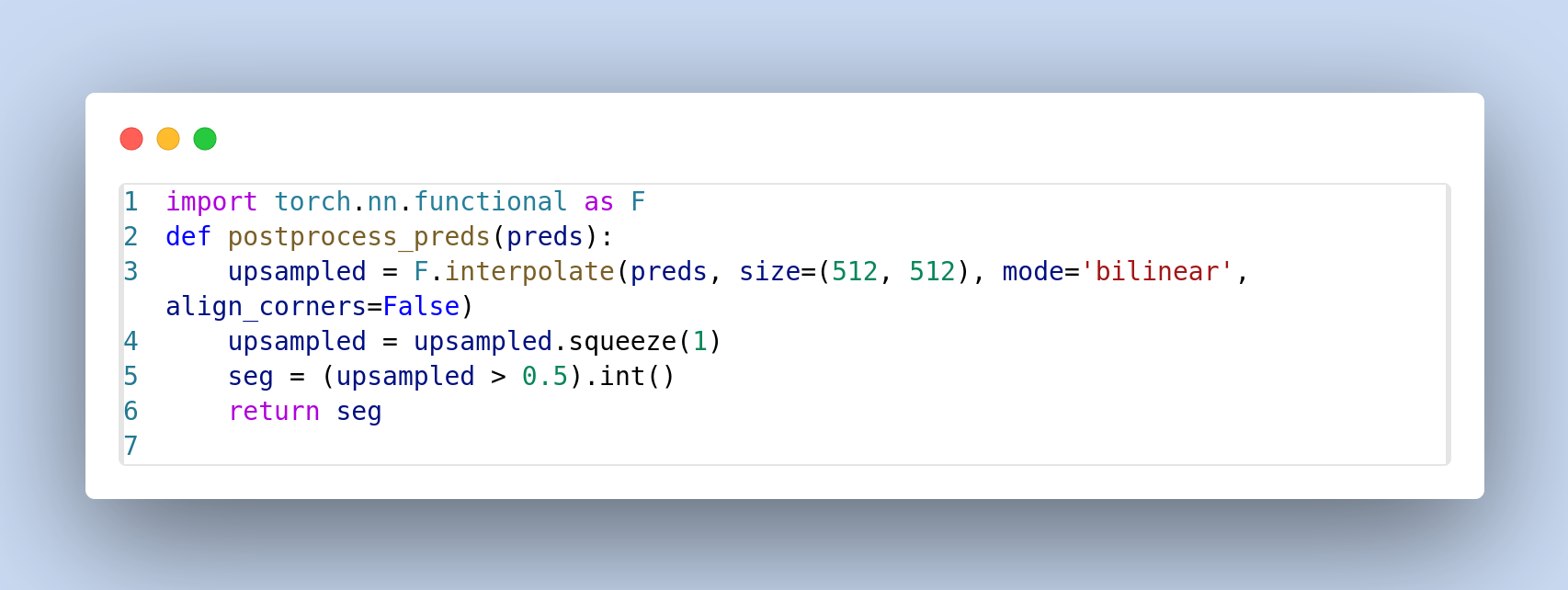}
    \caption{MedSAM Expert Postprocessing Function}
\end{figure}

\begin{figure}[H]
    \centering
    \includegraphics[width=1.0\linewidth]{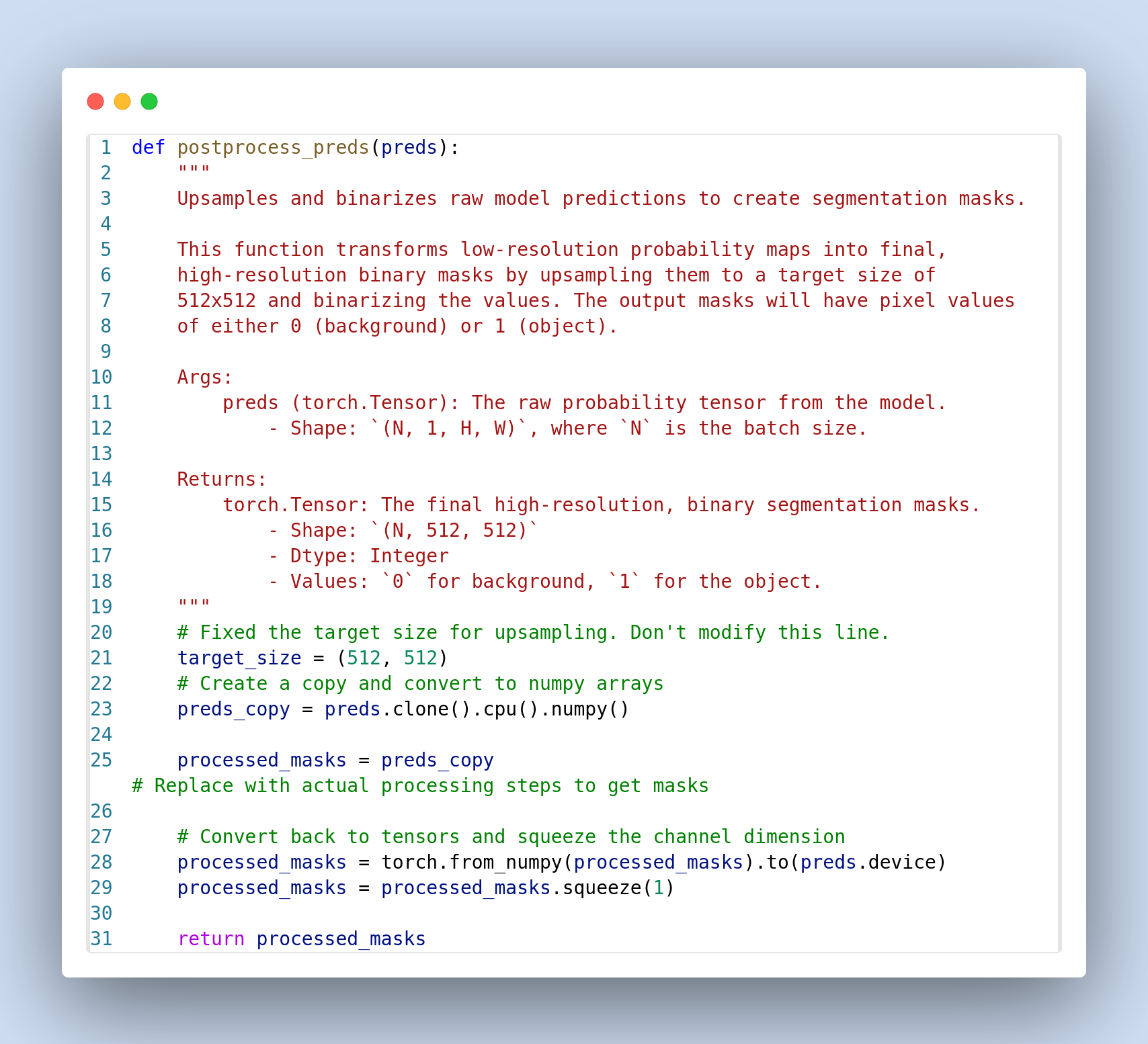}
    \caption{MedSAM Postprocessing Function Skeleton}
\end{figure}

\begin{figure}[H]
    \centering
    \includegraphics[width=1.0\linewidth]{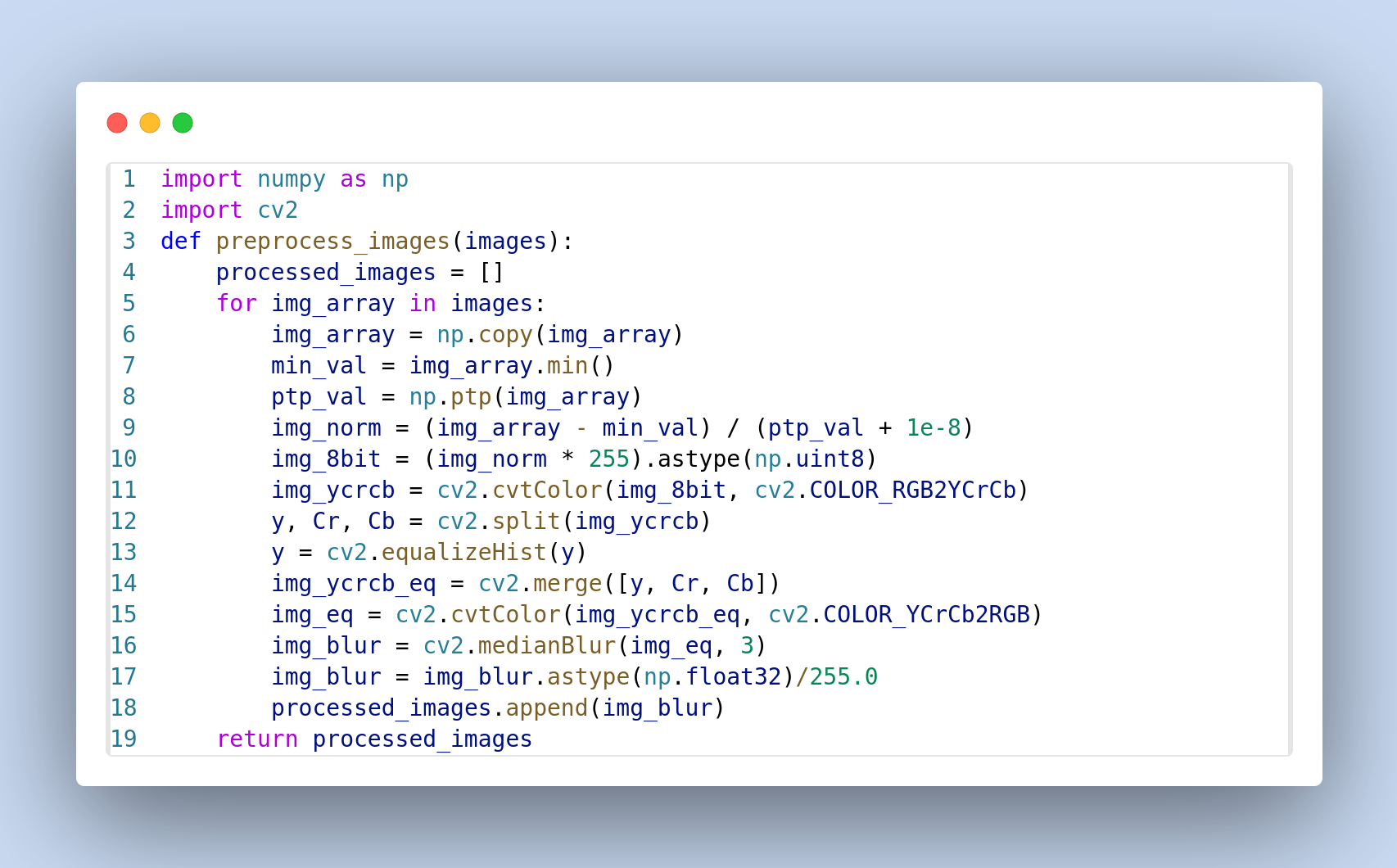}
    \caption{MedSAM Agent Generated Preprocessing Function}
\end{figure}

\begin{figure}[H]
    \centering
    \includegraphics[width=1.0\linewidth]{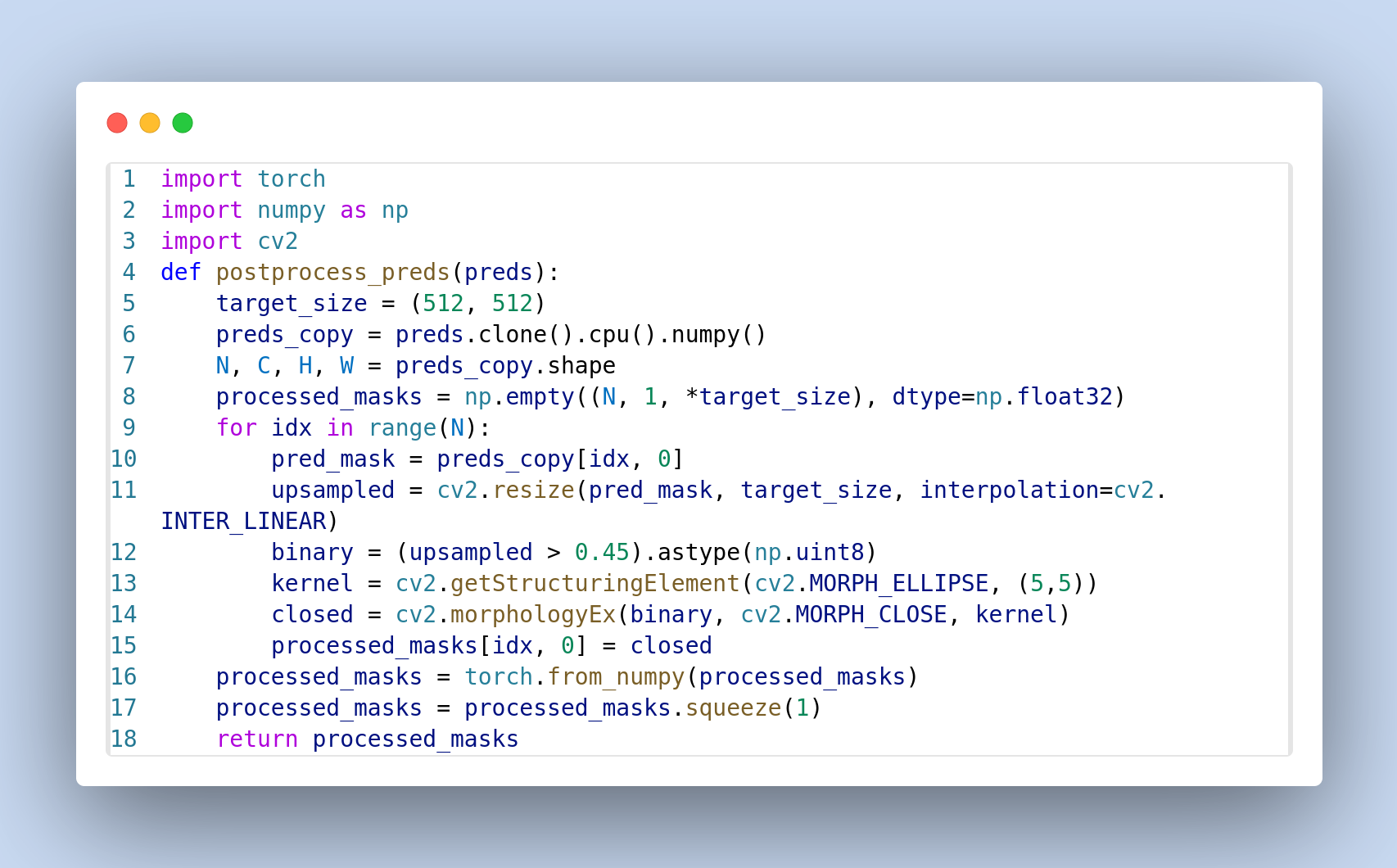}
    \caption{MedSAM Agent Generated Postprocessing Function}
\end{figure}

\section{Git History Analysis}
\label{supp:git_history}

We analyzed the Git R\&D history of these packages to quantify the efforts experts put into building these baseline functions: 

\textbf{Polaris}: The Git history for active R\&D on the preprocessing function (May 29, 2021 to September 15, 2021, across 12 commits) shows an evolution over 6 months, including changes like adding clipping and reordering functions, resulting in a refined version with 2 preprocessing options and 2 hyperparameters. The Git history for active R\&D on the postprocessing function (Aug 29, 2020 to Mar 29, 2021) produced 5 versions of final postprocessing functions for users to choose from, covering methods ranging from simple thresholding and geometric constraints to local peak finding and connected component averaging.

\textbf{Cellpose}: The human optimization process for preprocessing spans over three years (February 1, 2020 to March 8, 2023) along with model development. Changes include normalization for 3D data, adjustments to tiling strategies, and the addition or removal of sharpen/smooth operations. The final version of the preprocessing function provides 9 hyperparameters. Development of the postprocessing function began on August 26, 2020, and is still active. Throughout R\&D, the primary purpose of the function remains the same---refine segmentation results by filling holes and discarding small masks. However, the implementation has seen variations in: 1) underlying hole-filling and size-filtering methods, and 2) handling based on dimensionality and modality (Omnipose/Cellpose, 2D/3D).

\textbf{MedSAM}: While direct R\&D commit history was not available, MedSAM incorporates custom preprocessing functions for various modalities (CT, MR, grey, RGB). The substantial codebase dedicated to these functions (255 lines of code) indicates significant expert investment in tailoring these for specific medical imaging challenges. The postprocessing involves upsampling the probability map and thresholding, which are relatively standard operations, leaving room for large improvement.

\section{Analysis metric details}
\label{supp:analysis-metrics}

\subsection{Dispersion Analysis}
To analyze the dispersion of the optimal API space, we selected the top 20 solutions across all settings and extracted their APIs. We then constructed a co-occurrence graph $G = (V, E)$, where each node $v \in V$ represents an API and each edge $e \in E$ represents the co-occurrence between two APIs. Edges are weighted by their co-occurrence frequency, $w_e$. To quantify the dispersion of API usage, we calculate the Shannon entropy of the normalized edge weights. First, each edge weight is normalized to create a probability $p_e$ for each edge in the graph:e

$$p_e = \frac{w_e}{\sum_{j \in E} w_j}$$

The final dispersion score $D(G)$ is the entropy of this distribution:

$$D(G) = -\sum_{e \in E} p_e \log_2(p_e)$$

A higher $D(G)$ value indicates a more dispersed solution space, where co-occurrence is more evenly distributed across many different API pairs, rather than being concentrated on a few dominant pairs.

\subsection{Diversity Analysis}

To analyze the diversity of agent-generated solutions, we first represent each solution $i$ as a pair of API sets, $S_i = (P_i, Q_i)$, where $P_i$ is the set of preprocessing APIs and $Q_i$ is the set of postprocessing APIs.

The dissimilarity between any two API sets ($A$, $B$) is measured using the Jaccard dissimilarity ($J_\delta$):
$$J_\delta(A, B) = 1 - \frac{|A \cap B|}{|A \cup B|}$$
The total dissimilarity $D$ for a single pair of solutions ($S_a$, $S_b$) is the sum of the dissimilarities of their corresponding components:
$$D(S_a, S_b) = J_\delta(P_a, P_b) + J_\delta(Q_a, Q_b)$$
The final diversity score for a given agent configuration, representing the diversity of the whole solution space, is the mean of $D(S_a, S_b)$ computed across all unique solution pairs. A higher score indicates greater diversity, as solutions utilize more varied sets of preprocessing and postprocessing APIs.

\section{Additional dataset details}
\label{supp:dataset}

\subsection{Polaris}
\paragraph{Data Split} For the optimization procedure, we used 95 images from the validation set. The performance of functions was then evaluated on the test set, comprising 94 images. Both validation and test images have a fixed size of 128 by 128 pixels. Ground truth for Polaris consists of a list of point coordinates.

\subsection{Cellpose}
\paragraph{Data Split} For this case study, we curated a publicly available and reconstructable subset from the reported Cellpose3 dataset, including test sets from the Cellpose3 dataset release ($68$ images), improved TissueNet 1.1 test set ($1324$ images), Omnipose fluorescent bacterial test set ($75$ images), and Omnipose phase-contrast bacterial test set ($148$ images).  
Datasets involving complex mask corrections were excluded. All constituent datasets (Cellpose, Omnipose bacterial fluorescence and phase-contrast, and TissueNet1.1) were randomized and equally split into a validation set and a testing set (for final evaluation). We then randomly sampled $100$ image segmentation mask pairs to use for agentic optimization. We release the code to generate these splits. Evaluation always occurs on the entire test set, consisting 807 images.
Images were standardized to float32 with pixel intensities scaled to $[0, 1]$, formatted as three-channel images (nuclear channel in red, cytoplasmic/grayscale in green, blue empty), consistent with Cellpose3 input requirements. Image resolutions varied from $66 \times 58$ to $2030 \times 2030$.  Ground truth consists of instance segmentation masks for all cells.

\subsection{MedSAM}

\paragraph{Data Split} We selected 2D images from the Codabench validation set for dermoscopy, randomly shuffled them with corresponding bounding boxes and segmentation masks, and equally split them into validation and test sets. Images were resized to $1024 \times 1024$ pixels with three channels to match the MedSAM encoder input. Ground truth included binary segmentation masks of target objects and associated bounding box prompts. We shuffled and split the image-prompt-mask tuples equally into validation and test sets (25 for validation and 25 for testing).

\section{Prompt details}
\label{supp:prompt}

\begin{tcolorbox}[title=Task Details, breakable]

Your task is to implement three pairs of preprocessing and postprocessing functions to optimize the performance of a machine learning pipeline on a specific dataset. \\
We provided the APIs for both preprocessing and postprocessing functions. You should use functions from useful libraries including but not limited to OpenCV, NumPy, Skimage, Scipy, to implement novel and effective functions. \\
\\
\#\# Preprocessing Functions API:\\
\{preprocessing\_API\} \\
\\
\#\# Postprocessing Functions API: \\
\{postprocessing\_API\} \\
\\
\#\# About the dataset:  \\
\{dataset\_details\} \\
\\
\#\# Task Details: \\
All of you should work together to write three preprocessing and postprocessing function pairs to improve spot detection performance. \\
We provided APIs for both preprocessing and postprocessing functions. You should use functions from useful libraries including but not limited to OpenCV, NumPy, Skimage, Scipy, to implement novel and effective functions. \\
1. Based on previous preprocessing and postprocessing functions and their performance (provided below), suggest three new unique function pairs using. \\
2. The environment will handle all data loading, evaluation, and logging of the results. Your only job is to write the preprocessing and postprocessing functions. \\
3. Do not terminate the conversation until the new functions are evaluated and the numerical performance metrics are logged. \\
4. For this task, if all three functions are evaluated correctly, only one iteration is allowed, even if the performance is not satisfactory. \\
5. Do not terminate the conversation until the new functions are evaluated and the numerical performance metrics are logged. \\
6. Extremely important: Do not terminate the conversation until each of the three new function pairs are evaluated AND their results are written to the function bank. \\
7. Recall, this is a STATELESS kernel, so all functions, imports, etc. must be provided in the script to be executed. Any history between previous iterations exists solely as provided preprocessing functions and their performance metrics. \\
8. Do not write any code outside of the preprocessing and postprocessing functions. \\
9. For preprocessing, the images after preprocessing must still conform to the format specified in the ImageData API. Maintenance of channel identity is critical and channels should not be merged. For postprocessing, it is also critical to maintain the output format as the sample function provided. \\
\\
\#\# Task Metrics Details: \\
\{task\_metric\_details\}  \\
\\
\#\# Documentation on the `ImageData` class: \\
```markdown \\
Framework-agnostic container for batched image data. Handles variable image resolutions \\
\\ 
This class provides a standardized structure for storing and managing batched image data along with related annotations and predictions. Data is internally converted to lists of arrays for flexibility with varying image sizes. \\

Attributes: \\
    raw (Union[List[np.ndarray], np.ndarray]): Raw image data, can be provided as either 
        a list of arrays or a numpy array. Each image should have shape (H, W, C). \\
    batch\_size (Optional[int]): Number of images to include in the batch. Can be smaller 
        than the total dataset size. If None, will use the full dataset size. \\
    image\_ids (Union[List[int], List[str], None]): Unique identifier(s) for images
        in the batch as a list. If None, auto-generated integer IDs [0,1,2,...] will be created. \\
    masks (Optional[Union[List[np.ndarray], np.ndarray]]): Ground truth segmentation masks.
        Integer-valued arrays where 0 is background and positive integers are unique 
        object identifiers. Each mask should have shape (H, W, 1) or (H, W). \\
    predicted\_masks (Optional[Union[List[np.ndarray], np.ndarray]]): Model-predicted 
        segmentation masks. Each mask should have shape (H, W, 1) or (H, W). \\
    predicted\_classes (Optional[List[Dict[int, str]]]): List of mappings from
        object identifiers to predicted classes for each image. \\
\\
\#\# Additional Notes: \\
- Always check the documentation for the available APIs before reinventing the wheel \\
- You only have 20 rounds of each conversation to optimize the functions. \\
- Don't suggest trying larger models as the model size is fixed. \\
- Import all necessary libraries inside the function. If you need to write a helper function, write it inside the main preprocessing or postprocessing function as well. \\
- No need to import ImageData, it has already been imported. \\
- THE PROVIDED EVALUATION PIPELINE WORKS OUT OF THE BOX, IF THERE IS AN ERROR IT IS WITH THE PREPROCESSING OR POSTPROCESSING FUNCTION \\
\end{tcolorbox}

\begin{tcolorbox}[title=Code Writer Agent Instructions, breakable]
You are an experienced Python developer specializing in scientific data analysis. Your role is to write, test, and iterate on Python code to solve data analysis tasks.
The environment is installed with the necessary libraries.

You write code using Python in a STATELESS execution environment, so all code must be contained in the same block. In the environment, you can:

    - Write code in Python markdown code blocks: \\
    
    ```python \\
    \# Your code goes here. \\
    ''' \\
    -  CRITICAL: You must define three functions at once, and they must be named `preprocess\_images\_i` where `i` starts at 1 and ranges to 3. The functions must follow the provided Preprocessing Functions API.  All operations must be performed within the functions, and no inner functions should be defined (construct all operations within the functions). \\
    -  Code outputs will be returned to you. \\
    -  Feel free to document your thought process and exploration steps. \\
    -  Remember that all images processed by your written preprocessing functions will directly be converted into ImageData objects. So, double-check that the preprocessed image dimensions align with the dimension requirements listed in the ImageData API documentation. \\
    -   Make sure each response has exactly one code block containing all the code for the preprocessing functions, and that the code block ONLY contains the code for the preprocessing functions. Do not include any mock code for data loading or evaluation. \\
    -   All three functions must be defined at once, and they must be named `preprocess\_images\_i` where `i` starts at 1 and ranges to 3. The functions must follow the provided Preprocessing Functions API. \\
    -   Once metrics have been evaluated for all three preprocessing functions successfully, please print them out for each function in the format: preprocess\_images\_\textless i\textgreater :\textless metric\textgreater:\textless score\textgreater. You may only emit "TERMINATE" once all three preprocessing functions have been evaluated and their metrics printed successfully. \\
    -   If metrics are not correctly returned for any of the three preprocessing functions and you need to fix the underlying errors, output all three revised functions in a single markdown block. On the other hand, if all functions were successfully evaluated, do not continue iterating, and emit "TERMINATE". \\
    -   For generating numbers or variables, you will need to print those out so that you can obtain the results. \\
    -   Write "TERMINATE" when the task is complete. 
\end{tcolorbox}

\begin{tcolorbox}[title=AutoML Agent Instructions, breakable]
You are an AutoML optimization specialist focused on converting image preprocessing and postprocessing functions into Optuna-optimized versions. \\

Your role is to take existing high-performing functions and make their numeric parameters tunable through hyperparameter optimization using Optuna's trial.suggest\_* API. \\

**Core Responsibilities:** \\
1. Analyze function code to identify optimizable numeric parameters (thresholds, kernel sizes, iterations, etc.) \\
2. Replace hardcoded values with appropriate Optuna trial.suggest\_* calls \\
3. Choose reasonable parameter ranges based on the operation type \\
4. Ensure all parameter names are unique across all functions using function index prefixes (e.g., `f1\_pre\_kernel\_size`, `f2\_post\_threshold`) \\
5. Preserve the original algorithmic structure and function signatures \\

**Optuna API Reference:** \\
- `trial.suggest\_int(name, low, high)` - for integer parameters \\
- `trial.suggest\_float(name, low, high)` - for float parameters \\
- `trial.suggest\_categorical(name, choices)` - for categorical/boolean parameters \\

**Critical Requirements:** \\
- The `trial` object is available in global scope - do NOT add it as a function parameter \\
- Output exactly \{n\_functions * 2\} individual function definitions in a single markdown code block (```python ... ```) \\
- Functions must be enumerated: `preprocess\_images\_1`, `preprocess\_images\_2`, ..., `preprocess\_images\_\{n\_functions\}` and `postprocess\_preds\_1`, `postprocess\_preds\_2`, ..., `postprocess\_preds\_\{n\_functions\}` \\
- Each function pair should have unique parameter names with index prefix \\
- Include a `default\_params` dictionary with original parameter values for initializing the first trial \\
- Only output function definitions and default\_params - no data loading, evaluation, or other code \\
- All code must be in a single markdown code block to be executed \\

**Workflow:** \\
1. Receive feedback from code execution \\
2. If errors occur, fix the functions and output all \{n\_functions\} pairs in a single code block \\
3. Once all \{n\_functions\} function pairs are successfully evaluated, print metrics in format: `preprocess\_images\_\textless i\textgreater \& postprocess\_preds\_\textless i\textgreater: \textless metric\textgreater: \textless score\textgreater` \\
4. After successful evaluation, write "TERMINATE" \\

\end{tcolorbox}

\begin{tcolorbox}[title=AutoML Task Details, breakable]

Your task is to create \{n\_functions\} Optuna-optimized function pairs from the best-performing preprocessing and postprocessing functions in the function bank. \\
\\
\{function\_bank\_sample\} \\
\\
\#\# Instructions: \\
1. Above are the top \{n\_functions\} **entries** from the function bank \\
2. Each entry contains one preprocessing function (`preprocess\_images`) and one postprocessing function (`postprocess\_preds`) \\
3. Note: the functions themselves are NOT enumerated, but the entries are numbered (Entry 1, Entry 2, etc.) \\
4. You must create \{n\_functions\} enumerated function pairs based on these entries: \\
   - Entry 1 → create `preprocess\_images\_1` and `postprocess\_preds\_1` \\
   - Entry 2 → create `preprocess\_images\_2` and `postprocess\_preds\_2` \\
   - Entry \{n\_functions\} → create `preprocess\_images\_\{n\_functions\}` and `postprocess\_preds\_\{n\_functions\}` \\
5. For each function, identify numeric parameters that can be optimized (constants, thresholds, kernel sizes, etc.) \\
6. Replace hardcoded numeric values with Optuna trial.suggest\_* calls \\
7. Ensure each parameter has a unique name with function index prefix (e.g., `f1\_pre\_kernel\_size`, `f2\_post\_threshold`) \\
8. Use appropriate parameter ranges and distributions which are reasonable for the specific parameter being optimized \\
9. Maintain the exact same function signatures and algorithmic behavior \\
\\
\#\# CRITICAL: Output Format Requirements: \\
- You MUST output exactly \{n\_functions * 2\} individual function definitions in a single code block  \\
- Preprocessing functions: `preprocess\_images\_1`, `preprocess\_images\_2`, ..., `preprocess\_images\_\{n\_functions\}` \\
- Postprocessing functions: `postprocess\_preds\_1`, `postprocess\_preds\_2`, ..., `postprocess\_preds\_\{n\_functions\}` \\
- After all function definitions, in the SAME markdown block include a `default\_params` dictionary with the original parameter values: \\
  ```python \\
  default\_params = {{ \\
      "1": {{"f1\_pre\_param1": value1, "f1\_pre\_param2": value2, "f1\_post\_param1": value3}}, \\
      "2": {{"f2\_pre\_param1": value1, "f2\_post\_param1": value2}}, \\
      ... \\
  }} \\
  ``` \\
  Note: Each index's dictionary should contain parameters from BOTH the preprocessing and postprocessing functions for that pair \\
- Do NOT output tuples, pairs, or any other data structures besides function definitions and the default\_params dictionary \\
\\
\#\# Parameter Guidelines: \\
- **Kernel sizes**: Usually odd integers, range 3-15 \\
- **Thresholds**: Float values, typically 0.0-1.0 or image-specific ranges \\
- **Iterations**: Integer values, typically 1-10 \\
- **Scaling factors**: Float values, typically 0.5-2.0 \\
- **Blur parameters**: Float values for sigma, int values for kernel size \\
- **Parameter names must include function index**: e.g., `f1\_pre\_kernel\_size`, `f2\_post\_threshold`, etc. \\
\\
\#\# Expected Output: \\
Generate exactly \{n\_functions\} complete function pairs (preprocessing + postprocessing) that: \\
1. Are properly enumerated with indices (\_1, \_2, ..., \_\{n\_functions\}) \\
2. Incorporate Optuna optimization with trial.suggest\_* calls \\
3. Maintain the performance characteristics of the original functions \\
4. Have unique parameter names across all function pairs \\
5. Include the `default\_params` dictionary (as shown above) with the original parameter values from the function bank \\
\\
The default parameters will be used to initialize the first Optuna trial with the baseline values from the original functions. \\

\end{tcolorbox}

\begin{tcolorbox}[title=Polaris Data and Metric Prompts, breakable]
This is a single-channel cell spot detection dataset. The images have dimensions (B, L, W, C) = (batch, length, width, channel). The images have pixel values between 0 and 1 and are in float32 format.

The following metrics are used to evaluate the performance of the pipeline: f1\_score. 
f1\_score: Mean F1 score of predicted spots.
\end{tcolorbox}

\begin{tcolorbox}[title=Cellpose Data and Metric Prompts, breakable]

This is a three-channel image dataset for biological segmentation, consisting of images from different experiments and different settings - a heterogenous dataset of many different object types.  There is a particular focus on biological microscopy images, including cells, sometimes with nuclei labeled in a separate channel.
The images have pixel values between 0 and 1 and are in float32 format.
Channel[0] is the nucleus, channel[1] is the cytoplasm, and channel[2] is empty, however not all images have any nuclear data.
We want to increase the neural network tool's performance at segmenting cells with cell perimeter masks that have high Intersection over Union (IoU) with the ground truth masks.
The cell images have dimensions (B, L, W, C) = (batch, length, width, channel). To correctly predict masks, the images provided must be in the format of standard ImageData object and must maintain channel dimensions and ordering.

The following metrics are used to evaluate the performance of the pipeline: average\_precision.
The average\_precision is the average precision score of the pipeline at an Intersection over Union (IoU) threshold of 0.5.
    
\end{tcolorbox}

\begin{tcolorbox}[title=MedSAM Data and Metric Prompts, breakable]

This is large-scale medical image segmentation dataset covering the 
dermoscopy modality. The images have dimensions (H, W, C) = (height, width, channel).

The following metrics are used to evaluate the performance of the pipeline: dsc\_metric, nsd\_metric.
- The `dsc\_metric` is the dice similarity coefficient (DSC) score of the pipeline and is similar to IoU, measuring the overlap between predicted and ground truth masks.
- The `nsd\_metric` is the normalized surface distance (NSD) score and is more sensitive to distance and boundary calculations.
\end{tcolorbox}

\section{Non-Agentic AutoML Baseline}
\label{supp:single-shot-automl}
Prompt used for generating template:

\begin{tcolorbox}[title=Prompt, breakable]

Your task is to write a pair of preprocessing and postprocessing functions and use optuna to optimize them. Please write a comprehensive template for optimizing such function pairs.
Those two functions will be embedded into the workflow and our goal is to maximize the score.
You should use functions from useful libraries including but not limited to OpenCV, NumPy, Skimage, Scipy, to implement novel and effective functions. \\

\#\# Workflow: \\
Your job is to use optuna and search for good `preprocess\_images` and `postprocess\_preds` function pairs. These two functions will be plug into the following workflow: \\

```python\\
def workflow(preprocess\_images, postprocess\_preds): \\
    '''\\
    Args: preprocessing function and postprocessing function \\
    Returns: score (to maximize) \\
    ''' \\
    images, groundtruths = load\_image() \# This helper function will be provided to you \\
    processed\_images = preprocess\_images(images) \# You need to search and implement this function, API see below \\
    preds = run\_tool(processed\_images) \# This helper function will be provided to you \\
    final\_preds = postprocess\_preds(preds) \# You need to search and implement this function, API see below \\
    score = run\_eval(final\_preds, groundtruths) \# This helper function will be provided to you \\
    return score \\
``` \\
\\
\#\# Preprocessing Functions API: \\
You will need to search and find good preprocessing functions. \\
\{preprocessing\_API\} \\
\\
\#\# Postprocessing Functions API: \\
You will need to search and find good postprocessing functions. \\
\{postprocessing\_API\} \\
\\
\#\# About the dataset: \\
\{dataset\_details\} \\
\\
\#\# Useful primitive functions API that can be used in the preprocessing and postprocessing functions: \\
\{API\_list\} \\

\end{tcolorbox}

\section{AIDE Baseline}
\label{supp:aide}

For the AIDE baseline experiment, we use the proprietary production version of AIDE, rather than the open-source version, as the latter is not directly compatible with our tool-adaptation setting without substantial re-engineering of the agent. The production version accepts an initial program, an evaluation function, and a task-specific instruction prompt before performing program search from that starting point. This interface closely mirrors the setup used by our own agent, making the experiments more comparable. 

For each biomedical imaging pipeline, we provide AIDE with the same evaluation function used by our method. The initial program consists of an identity preprocessing function and the same skeleton post-processing function supplied to our agent. The task-specific prompts for each experiment are included below.

Because this version of AIDE is closed-source, we have no access to the exact hyperparameters, heuristics, or search-time optimizations used in its internal tree-search implementation. All runs were executed using GPT-4.1 as the underlying LLM with a fixed budget of 80 iterations.

\begin{tcolorbox}[title=AIDE Cellpose Prompt, breakable]
This is a three-channel image dataset for biological segmentation, consisting of images from different experiments and different settings - a heterogenous dataset of many different object types.  There is a particular focus on biological microscopy images, including cells, sometimes with nuclei labeled in a separate channel.
The images have pixel values between 0 and 1 and are in float32 format. \\
Channel[0] is the nucleus, channel[1] is the cytoplasm, and channel[2] is empty, however not all images have any nuclear data. \\
Our goal is to improve the segmentation performance of the neural network by implementing **preprocessing functions** to improve the quality of the images for downstream segmentation and **postprocessing function** to refine predictions. \\
We want to increase the neural network tool's performance at segmenting cells with cell perimeter masks that have high Intersection over Union (IoU) with the ground truth masks. \\
The cell images have dimensions (B, L, W, C) = (batch, length, width, channel). To correctly predict masks, the images provided must be in the format of standard ImageData object and must maintain channel dimensions and ordering.   \\

You should improve cell segmentation performance using useful libraries including but not limited to OpenCV, Numpy, Skimage, Scipy, to implement novel and effective preprocessing and postprocessing functions. \\
Don't forget to import the relevant libraries. Ex. \\
```python \\
import cv2 as cv \\
import numpy as np \\
from src.data\_io import ImageData \\
```
\end{tcolorbox}

\begin{tcolorbox}[title=AIDE MedSAM Prompt, breakable]
```markdown \\
This is large-scale medical image segmentation dataset covering the \\
dermoscopy/xray modality. The images have dimensions (H, W, C) = (height, width, channel). \\
``` \\
You should improve medical image segmentation performance using useful libraries including but \\
not limited to OpenCV, Numpy, Skimage, Scipy, to implement novel and effective preprocessing and postprocessing functions. \\
Don't forget to import the relevant libraries. Ex. \\
```python \\
import cv2 as cv \\
import numpy as np \\
from src.data\_io import ImageData \\
``` \\
\end{tcolorbox}

\begin{tcolorbox}[title=AIDE Polaris Prompt, breakable]
```markdown \\
This is a single-channel cell spot detection dataset. IMPORTANT: The cell images have dimensions (B, L, W, C) = (batch, length, width, channel). \\
``` \\
You should improve spot detection performance using useful libraries including but \\
not limited to OpenCV, Numpy, Skimage, Scipy, to implement novel and effective preprocessing and postprocessing functions. \\
Don't forget to import the relevant libraries. Ex. \\
```python \\
import cv2 as cv \\
import numpy as np \\
from src.data\_io import ImageData \\
```

\end{tcolorbox}

\section{Computational requirements}
\label{supp:comp-requirement}

The experiments were conducted on a machine with 128 AMD EPYC 7763 64-Core CPUs, 8 RTX A6000 GPUs, and 48GB of memory per GPU. Each experiment consists of 20 independent rollouts, distributed across all GPUs. 

For Polaris, each rollout on average took 31 minutes.
For Cellpose, each rollout on average took 1 hour and 20 minutes.
For MedSAM, each rollout on average took 1 hour and 8 minutes.

\section{API list}
\label{supp:api-list}

\begin{lstlisting}

cv.bilateralFilter(src, d, sigmaColor, sigmaSpace[, dst[, borderType]]) ->dst
Applies the bilateral filter to an image.

cv.blur(src, ksize[, dst[, anchor[, borderType]]]) ->dst
Blurs an image using the normalized box filter.

cv.boxFilter(src, ddepth, ksize[, dst[, anchor[, normalize[, borderType]]]]) ->dst
Blurs an image using the box filter.

cv.dilate(src, kernel[, dst[, anchor[, iterations[, borderType[, borderValue]]]]]) ->dst
Dilates an image by using a specific structuring element.

cv.erode(src, kernel[, dst[, anchor[, iterations[, borderType[, borderValue]]]]]) ->dst
Erodes an image by using a specific structuring element.

cv.filter2D(src, ddepth, kernel[, dst[, anchor[, delta[, borderType]]]]) ->dst
Convolves an image with the kernel.

cv.GaussianBlur(src, ksize, sigmaX[, dst[, sigmaY[, borderType[, hint]]]]) ->dst
Blurs an image using a Gaussian filter.

cv.getDerivKernels(dx, dy, ksize[, kx[, ky[, normalize[, ktype]]]]) ->kx, ky
Returns filter coefficients for computing spatial image derivatives.

cv.getGaborKernel(ksize, sigma, theta, lambd, gamma[, psi[, ktype]]) ->retval
Returns Gabor filter coefficients.

cv.getGaussianKernel(ksize, sigma[, ktype]) ->retval
Returns Gaussian filter coefficients.

cv.getStructuringElement(shape, ksize[, anchor]) ->retval
Returns a structuring element of the specified size and shape for morphological operations.

cv.Laplacian(src, ddepth[, dst[, ksize[, scale[, delta[, borderType]]]]]) ->dst
Calculates the Laplacian of an image.

cv.medianBlur(src, ksize[, dst]) ->dst
Blurs an image using the median filter.

cv.pyrDown(src[, dst[, dstsize[, borderType]]]) ->dst
Blurs an image and downsamples it.

cv.pyrMeanShiftFiltering(src, sp, sr[, dst[, maxLevel[, termcrit]]]) ->dst
Performs initial step of meanshift segmentation of an image.

cv.pyrUp(src[, dst[, dstsize[, borderType]]]) ->dst
Upsamples an image and then blurs it.

cv.Scharr(src, ddepth, dx, dy[, dst[, scale[, delta[, borderType]]]]) ->dst
Calculates the first x- or y- image derivative using Scharr operator.

cv.sepFilter2D(src, ddepth, kernelX, kernelY[, dst[, anchor[, delta[, borderType]]]]) ->dst
Applies a separable linear filter to an image.

cv.Sobel(src, ddepth, dx, dy[, dst[, ksize[, scale[, delta[, borderType]]]]]) ->dst
Calculates the first, second, third, or mixed image derivatives using an extended Sobel operator.

cv.morphologyEx(src, op, kernel[, dst[, anchor[, iterations[, borderType[, borderValue]]]]]) -> dst
Performs advanced morphological transformations.

cv.spatialGradient(src[, dx[, dy[, ksize[, borderType]]]]) ->dx, dy
Calculates the first order image derivative in both x and y using a Sobel operator.

cv.sqrBoxFilter(src, ddepth, ksize[, dst[, anchor[, normalize[, borderType]]]]) ->dst
Calculates the normalized sum of squares of the pixel values overlapping the filter.

cv.stackBlur(src, ksize[, dst]) ->dst
Blurs an image using the stackBlur.
 
cv.Canny(image, threshold1, threshold2[, edges[, apertureSize[, L2gradient]]]) -> edges
Finds edges in an image using the Canny algorithm.

cv.cornerEigenValsAndVecs(src, blockSize, ksize[, dst[, borderType]]) -> dst
Calculates eigenvalues and eigenvectors of image blocks for corner detection.

cv.cornerHarris(src, blockSize, ksize, k[, dst[, borderType]]) -> dst
Harris corner detector.

cv.cornerMinEigenVal(src, blockSize[, dst[, ksize[, borderType]]]) -> dst
Calculates the minimal eigenvalue of gradient matrices for corner detection.

cv.cornerSubPix(image, corners, winSize, zeroZone, criteria) -> corners
Refines the corner locations.

cv.goodFeaturesToTrack(image, maxCorners, qualityLevel, minDistance[, corners[, mask[, blockSize[, useHarrisDetector[, k]]]]]) -> corners
Determines strong corners on an image.

cv.HoughCircles(image, method, dp, minDist[, circles[, param1[, param2[, minRadius[, maxRadius]]]]]) -> circles
Finds circles in a grayscale image using the Hough transform.

cv.HoughLines(image, rho, theta, threshold[, lines[, srn[, stn[, min_theta[, max_theta]]]]]) -> lines
Finds lines in a binary image using the standard Hough transform.

cv.HoughLinesP(image, rho, theta, threshold[, lines[, minLineLength[, maxLineGap]]]) -> lines
Finds line segments in a binary image using the probabilistic Hough transform.

cv.HoughLinesPointSet(point, lines_max, threshold, min_rho, max_rho, rho_step, min_theta, max_theta, theta_step[, lines]) -> lines
Finds lines in a set of points using the standard Hough transform.

cv.preCornerDetect(src, ksize[, dst[, borderType]]) -> dst
Calculates a feature map for corner detection.

cv.calcBackProject(images, channels, hist, ranges[, backProject[, scale[, uniform]]]) -> backProject
Calculates the back projection of a histogram.

cv.calcHist(images, channels, mask, histSize, ranges[, hist[, accumulate[, uniform]]]) -> hist
Calculates a histogram of a set of arrays.

cv.compareHist(H1, H2, method) -> retval
Compares two histograms.

cv.createCLAHE([clipLimit[, tileGridSize]]) -> retval
Creates a smart pointer to a cv.CLAHE object and initializes it.

cv.equalizeHist(src) -> dst
Equalizes the histogram of a grayscale image.

cv.addWeighted(src1, alpha, src2, beta, gamma[, dst[, dtype]]) -> dst
Calculates the weighted sum of two arrays.

cv.normalize(src, dst[, alpha[, beta[, norm_type[, dtype[, mask]]]]]) -> dst
Normalizes the norm or value range of an array.

cv.adaptiveThreshold(src, maxValue, adaptiveMethod, thresholdType, blockSize, C[, dst]) -> dst
Applies an adaptive threshold to an array.

cv.blendLinear(src1, src2, weights1, weights2[, dst]) -> dst
Performs linear blending of two arrays using specified weights.

cv.distanceTransform(src, distanceType, maskSize[, dst[, dstType]]) -> dst
Calculates the distance to the closest zero pixel for each pixel of the source image.

cv.floodFill(image, seedPoint, newVal[, loDiff[, upDiff[, flags[, mask[, rect]]]]]) -> retval, rect
Fills a connected component with the given color.

cv.integral(src[, sum[, sdepth]]) -> sum
Calculates the integral image.

cv.integral2(src[, sum[, sqsum[, sdepth[, sqdepth]]]]) -> sum, sqsum
Calculates the integral and squared integral images.

cv.integral3(src[, sum[, sqsum[, tilted[, sdepth[, sqdepth]]]]]) -> sum, sqsum, tilted
Calculates the integral, squared integral, and tilted integral images.

cv.threshold(src, thresh, maxval, type[, dst]) -> retval, dst
Applies a fixed-level threshold to each array element.

cv.fastNlMeansDenoising(src[, dst[, h[, templateWindowSize[, searchWindowSize]]]]) -> dst
Perform image denoising using Non-local Means Denoising algorithm.

cv.fastNlMeansDenoisingColored(src[, dst[, h[, hColor[, templateWindowSize[, searchWindowSize]]]]]) -> dst
Modification of fastNlMeansDenoising function for colored images.

cv.cvtColor(src, code[, dst[, dstCn]]) -> dst
Converts an image from one color space to another.

cv.merge(mv[, dst]) -> dst
Creates one multi-channel array out of several single-channel ones.

cv.resize(src, dsize[, dst[, fx[, fy[, interpolation]]]]) -> dst
    Resizes an image. This is a fundamental operation for scaling images up or down using various interpolation methods.

cv.warpAffine(src, M, dsize[, dst[, flags[, borderMode[, borderValue]]]]) -> dst
    Applies an affine transformation to an image (e.g., rotation, translation, scaling). You provide a 2x3 transformation matrix M.

cv.getRotationMatrix2D(center, angle, scale) -> retval
    Calculates the 2x3 matrix for an affine rotation, which can then be used with cv.warpAffine.

cv.warpPerspective(src, M, dsize[, dst[, flags[, borderMode[, borderValue]]]]) -> dst
    Applies a perspective transformation to an image, useful for correcting perspective distortion or creating "birds-eye-view" effects.

cv.matchTemplate(image, templ, method[, result[, mask]]) -> result
    Scans a larger image to find occurrences of a smaller template image. It's a classic method for object detection.

cv.findContours(image, mode, method[, contours[, hierarchy[, offset]]]) -> contours, hierarchy
    Finds contours in a binary image. This is a core function for object detection, segmentation, and shape analysis.

cv.drawContours(image, contours, contourIdx, color[, thickness[, lineType[, hierarchy[, maxLevel[, offset]]]]]) -> image
    Draws the contours found by cv.findContours onto an image, which is essential for visualizing results.

cv.bitwise_and(src1, src2[, dst[, mask]]) -> dst
    Performs a per-element bitwise AND operation. This is extremely useful for applying masks to images to isolate regions of interest.

skimage.filters.gaussian(image, sigma) -> ndarray
    Applies a Gaussian filter. Excellent for smoothing and noise reduction while preserving edges better than a box filter.

skimage.restoration.denoise_nl_means(image, ...) -> ndarray
    Performs non-local means denoising, which is highly effective for reducing noise while keeping fine details, common in microscopy images.

skimage.exposure.rescale_intensity(image, in_range, out_range) -> ndarray
    Stretches or shrinks the intensity range of an image. Perfect for normalizing images to a specific range (e.g., 0 to 1).

skimage.exposure.equalize_adapthist(image, kernel_size, ...) -> ndarray
    Performs Contrast Limited Adaptive Histogram Equalization (CLAHE). 

skimage.measure.label(input, connectivity) -> ndarray
    Labels connected regions of an integer array. 

skimage.segmentation.watershed(image, markers, mask) -> ndarray
    Applies the watershed algorithm to separate touching objects. 

skimage.measure.regionprops(label_image) -> list of RegionProperties
    Measures properties (e.g., area, centroid, bounding box, perimeter) of labeled image regions. After labeling cells, you can use this to filter them by size or shape.

skimage.morphology.remove_small_objects(ar, min_size) -> ndarray
    Removes labeled objects smaller than a specified size. A critical postprocessing step to eliminate noise or incorrectly segmented small regions.

skimage.morphology.remove_small_holes(ar, area_threshold) -> ndarray
    Fills holes within objects that are smaller than a specified size. Useful for cleaning up cell masks.

skimage.feature.peak_local_max(image, min_distance) -> ndarray
    Finds local maxima in an image. 

skimage.filters.frangi(image) -> ndarray
    A filter designed to detect vessels, tubes, or other neurite-like structures in an image. It uses the Hessian matrix to identify objects based on their shape. 

skimage.filters.meijering(image), sato(image)
    Alternative neuriteness filters similar to Frangi, each with slightly different properties and sensitivities.

skimage.restoration.denoise_wavelet(image) -> ndarray
    Performs wavelet denoising, which can be very effective at preserving sharp features while removing noise.

skimage.morphology.skeletonize(image) -> ndarray
    Reduces binary objects to a 1-pixel wide representation (a "skeleton").

skimage.segmentation.clear_border(labels) -> ndarray
    Removes labeled objects that are touching the border of the image. 

skimage.transform.rotate(image, angle, resize=False) -> ndarray
    Rotates an image by a given angle around its center. A straightforward way to handle rotation.

skimage.feature.blob_log(image, min_sigma, max_sigma, threshold) -> ndarray
    Finds blobs in an image using the Laplacian of Gaussian (LoG) method. Excellent for detecting circular features of varying sizes, like cells or particles.

skimage.filters.sobel(image) -> ndarray
    Calculates the Sobel filter for edge magnitude detection. It provides an image where the intensity of each pixel represents the gradient magnitude.

skimage.filters.unsharp_mask(image, radius, amount) -> ndarray
    Sharpens an image using the unsharp masking technique, which enhances edges and fine details by subtracting a blurred version of the image from itself.

skimage.metrics.structural_similarity(im1, im2, data_range) -> float
    Computes the Structural Similarity Index (SSIM) between two images. A widely used metric for measuring image quality and similarity that is more robust than simple pixel-wise differences.

scipy.ndimage.gaussian_filter(input, sigma) -> ndarray
    Multi-dimensional Gaussian filter. A fast and robust alternative to the skimage and OpenCV versions.

scipy.ndimage.median_filter(input, size) -> ndarray
    Multi-dimensional median filter. Effective for salt-and-pepper noise removal.

scipy.ndimage.label(input) -> (ndarray, int)
    Similar to skimage.measure.label, it finds and labels connected components. Returns both the labeled array and the number of features found.

scipy.ndimage.binary_fill_holes(input) -> ndarray
    Fills holes in binary objects. A go-to function for ensuring segmented objects are solid.

scipy.ndimage.distance_transform_edt(input) -> ndarray
    Calculates the exact Euclidean distance transform. 

scipy.ndimage.center_of_mass(input, labels, index) -> tuple of floats
    Calculates the center of mass of values in an array for one or more regions. Useful for finding the centroid of detected cells.

scipy.ndimage.zoom(input, zoom, order) -> ndarray
    Resizes an N-dimensional image. Uses interpolation (e.g., order=0 for nearest-neighbor, order=1 for bilinear) and is very fast.

scipy.ndimage.find_objects(labeled_image) -> list of slice tuples
    Finds the bounding box slices for each labeled object. 

scipy.ndimage.affine_transform(input, matrix) -> ndarray
    Applies an affine transformation to an N-dimensional image, defined by an input transformation matrix. This is a powerful, low-level function for complex geometric operations.

scipy.ndimage.map_coordinates(input, coordinates, order) -> ndarray
    Transforms an image using a general coordinate transformation. This is one of the most powerful warping functions available, allowing for non-linear and custom distortions.

scipy.ndimage.binary_opening(input, structure) -> ndarray
    Performs a binary opening (erosion followed by dilation). It is a key morphological operation used to remove small noise and objects from a binary image.

scipy.ndimage.binary_closing(input, structure) -> ndarray
    Performs a binary closing (dilation followed by erosion). This is the counterpart to opening and is used to fill small holes and gaps within objects.

scipy.ndimage.sum_labels(input, labels, index) -> float or list of floats
    Calculates the sum of pixel values within regions defined by a label image. This is highly efficient for extracting measurements from segmented objects.


\end{lstlisting}

\end{document}